\documentclass[10pt,twocolumn,letterpaper]{article}

\usepackage{iccv}              
\usepackage{booktabs}  
\usepackage{xcolor}    
\usepackage{colortbl}  
\usepackage{graphicx}  
\usepackage{url}
\usepackage{float}
\usepackage{array}     
\usepackage{dsfont}
\usepackage{enumitem} 
\usepackage{xcolor}  
\usepackage{graphicx}
\usepackage{amsmath}
\usepackage{amssymb}
\usepackage{tabularx}
\usepackage{threeparttable}
\usepackage{booktabs} 
\usepackage{booktabs}
\usepackage{booktabs}  
\usepackage{xcolor}    
\usepackage{colortbl}  
\usepackage{graphicx}  
\usepackage{url}
\usepackage{float}
\usepackage{array} 
\usepackage{enumitem}
\usepackage{url}
\usepackage{pifont}
\newcommand{\cmark}{\ding{51}} 
\newcommand{\xmark}{\ding{55}}
\definecolor{mypink2}{rgb}{.99,.96,.98}
\definecolor{mypink1}{rgb}{.99,.93,.98}
\definecolor{mypink}{rgb}{.99,.90,.98}
\definecolor{mygray}{rgb}{.95,.95,.95}
\definecolor{iccvblue}{rgb}{0.21,0.49,0.74}
\usepackage[pagebackref,breaklinks,colorlinks,allcolors=iccvblue]{hyperref}

\def\paperID{12814} 
\def\confName{ICCV}
\def\confYear{2025}

\title{Multi-Cache Enhanced Prototype Learning for Test-Time Generalization of Vision-Language Models}

\author{
    Xinyu Chen\textsuperscript{1,2,*} \quad
    Haotian Zhai\textsuperscript{2,3,*} \quad
    Can Zhang\textsuperscript{2} \quad
    Xiupeng Shi\textsuperscript{1,†} \quad
    Ruirui Li\textsuperscript{2,†} \\
    \vspace{2mm} 
    \textsuperscript{1}Shanghai University \quad
    \textsuperscript{2}Beijing University of Chemical Technology \quad
    \textsuperscript{3}University of Minnesota\quad \\
    {\tt\small \{xinyuchen,sxp\}@shu.edu.cn;haoti002@umn.edu;\{alexlessend,ilydouble\}@gmail.com}
}

\begin{document}
\maketitle
{\let\thefootnote\relax\footnotetext{\textsuperscript{*}Equal contribution. \textsuperscript{†}Corresponding author.}}
\begin{abstract}
In zero-shot setting, test-time adaptation adjusts pre-trained models using unlabeled data from the test phase to enhance performance on unknown test distributions. Existing cache-enhanced TTA methods rely on a low-entropy criterion to select samples for prototype construction, assuming intra-class compactness. However, low-entropy samples may be unreliable under distribution shifts, and the resulting prototypes may not ensure compact intra-class distributions. This study identifies a positive correlation between cache-enhanced performance and intra-class compactness. Based on this observation, we propose a Multi-Cache enhanced Prototype-based Test-Time Adaptation (MCP) featuring three caches: an entropy cache for initializing prototype representations with low-entropy samples, an align cache for integrating visual and textual information to achieve compact intra-class distributions, and a negative cache for prediction calibration using high-entropy samples. We further developed MCP++, a framework incorporating cross-modal prototype alignment and residual learning, introducing prototype residual fine-tuning. Comparative and ablation experiments across 15 downstream tasks demonstrate that the proposed method and framework achieve state-of-the-art generalization performance.{Project Page available at: \href{https://zhaihaotian.github.io/MCP-ICCV25/}{https://zhaihaotian.github.io/MCP-ICCV25/}}
\end{abstract}    
\vspace{-5pt}
\section{Introduction}
\label{sec:intro}

Vision-language pre-trained models, such as  CLIP \cite{CLIP}, ALIGN \cite{Jia2021ScalingUV}, have demonstrated strong zero-shot learning capabilities across various downstream tasks and have become a research focus in recent years. However, in real-world scenarios, differences in data collection environments or hardware devices often lead to distribution shifts between pre-training data and test data. This necessitates adaptive adjustments \cite{CLIP,Liang2023ACS,Wang2018DeepVD} during model deployment to maintain robust performance. To address this, Test-Time Adaptation (TTA) \cite{Zhang2021MEMOTT,Wang2020FullyTA, Lee2024EntropyIN} methods have been proposed, aiming to enable models to quickly adapt to new data distributions during the inference phase, effectively mitigating 
the negative impact of distribution shifts on model performance.

\begin{figure}[t]
  \centering
   \includegraphics[width=0.9\linewidth]{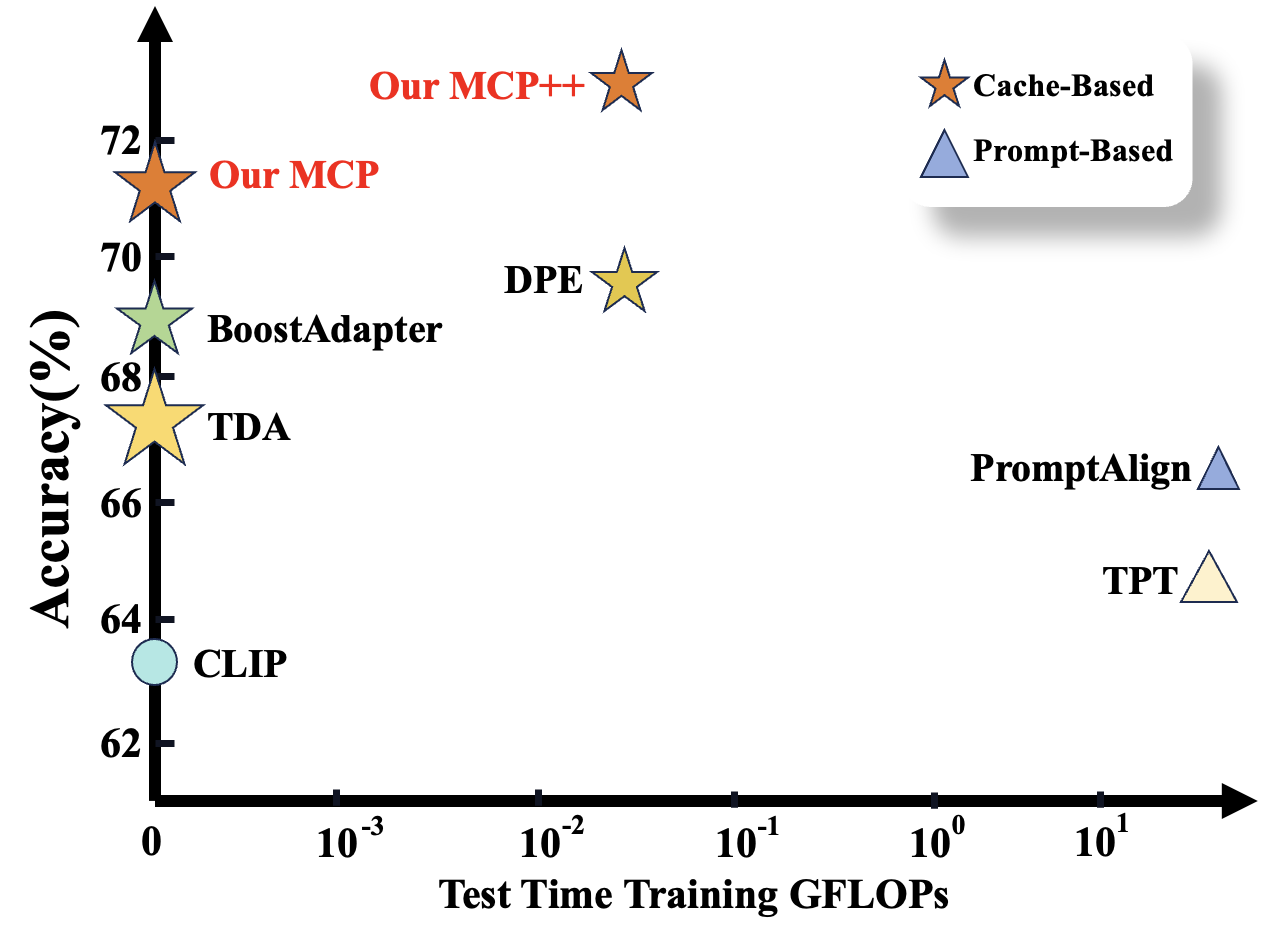}
   \caption{Illustration of the average classification accuracy, test-time training GFLOPs, and FPS for different methods on cross-dataset classification tasks. For prompt-based and cache-based methods, the icon sizes denote the FPS values.}
   \vspace{-10pt}
   \label{fig:acc}
\end{figure}

Recent studies have introduced TTA strategies for adapting vision-language models to downstream tasks \cite{TPT, PromptAlign, TPS, TDA, Zhang2024BoostAdapterIV, Zhang2024DualMN,hatem2023point,sun2025point}. Test-Time Prompt Tuning (TPT) \cite{TPT} pioneered this approach by learning domain-specific prompts from test data, enforcing prediction consistency across high-confidence augmented views of each test sample to dynamically refine the prompts. While this method effectively enhances CLIP's generalization capabilities, prompt-based approaches remain computationally intensive. As a result, cache-enhanced methods \cite{Zhang2024DualMN, Zhang2024BoostAdapterIV, TDA, DPE}, such as Test-Time Dynamic Adaptation (TDA) \cite{TDA}, have gained increasing attention. These methods build dynamic caches by selecting a limited number of reliable samples based on entropy criteria, thereby improving model adaptability to data from the test domain.

Existing cache-enhanced methods are based on the critical assumption that samples from the same class exhibit compactness in the feature space, ensuring the high representativeness of cached samples through low-entropy selection \cite{TDA,Zhang2024BoostAdapterIV,Zhang2024DualMN,DPE}. However, this assumption often breaks down under domain shift conditions: even low-entropy samples may be affected by spurious correlations, making them unreliable \cite{hu2025beyond,Lee2024EntropyIN,han2025ranked}, while the distributions of different classes can become overly dispersed, resulting in insufficient intra-class compactness. This lack of compactness hinders the construction of representative class prototypes. To observe these phenomena, we conduct a quantitative analysis of the relationship between cache-enhanced performance and intra-class compactness, aiming to uncover the sources of both the benefits and limitations of caching mechanisms. Our goal is to provide a solid theoretical foundation for improving these methods.

\begin{figure}[t]
  \centering
   \includegraphics[width=0.95\linewidth]{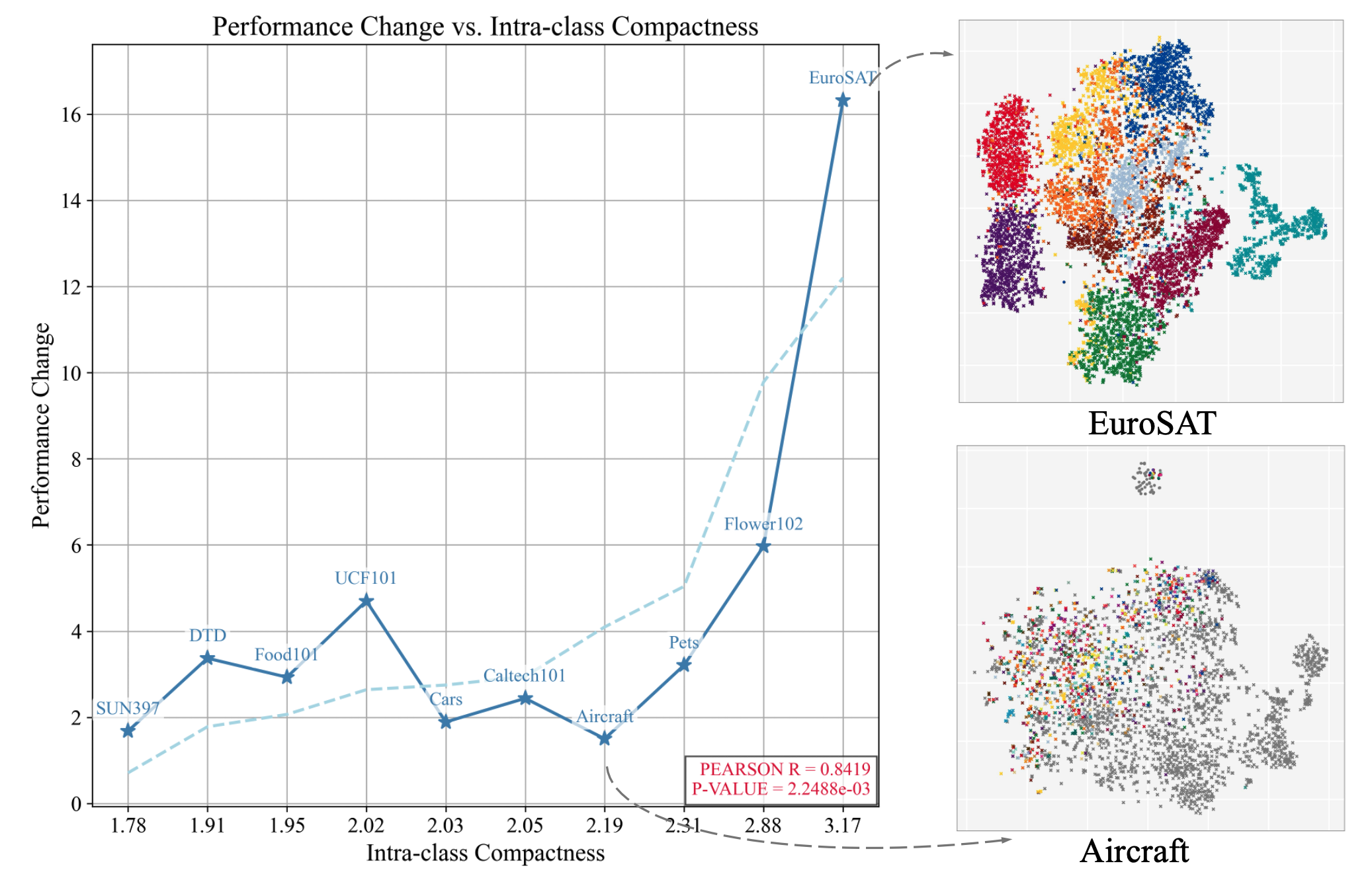}
   \caption{The $x$-axis represents the compactness of test data(defined as the inverse of the average distance between each sample and its class center), and the $y$-axis represents the accuracy improvement of TDA \cite{TDA} relative to zero-shot CLIP. A positive correlation is observed on the curve and two t-SNE visualizations on EuroSAT \cite{helber2019eurosat} and Aircraft \cite{aircraft} datasets.}
   \vspace{-10pt}
   \label{fig:intro}
\end{figure}

As shown in Fig.\ref{fig:intro}, a positive correlation is observed between the performance improvement of TDA over zero-shot CLIP and the compactness of cached samples per class. Specifically, the relative accuracy improvement is computed across test datasets, while compactness is defined as the reciprocal of the average distance between cached samples and their respective class centroids. The results indicate that TDA achieves notable performance gains on datasets with compact class distributions (e.g., EuroSAT), whereas the improvements are more limited on datasets with dispersed distributions (e.g., Aircraft). A Pearson correlation analysis ($r > 0.8, p = 2.25\times10^{-3}$) further validates this positive relationship, highlighting the critical role of intra-class compactness in enhancing cache structures.

Based on these insights, we propose a Multi-Cache enhanced Prototype-based Test-Time Adaptation method (MCP). Specifically, we first construct the \textbf{Entropy Cache} by selecting the lowest entropy samples for each class and dynamically updating it to provide initial prototype representations. However, as emphasized in our previous analysis, due to distribution shifts in test data, low-entropy samples do not always ensure compact visual feature distributions, which can blur class boundaries and limit the performance gains from the cache. To address this, we further integrate visual cache features and textual semantic information to optimize prototypes, forming the \textbf{Align Cache}. The align cache not only requires samples to have sufficiently low entropy but also evaluates their distance to the prototype center, prioritizing those that promote compact intra-class distributions. Additionally, high-entropy samples often contain critical information that helps suppress incorrect predictions. To fully leverage this, we construct the \textbf{Negative Cache} to provide negative references for the model, enabling more accurate calibration of prediction probabilities.

To optimize visual and textual prototypes further, we propose a cross-modal prototype alignment residual learning mechanism and develop a comprehensive multi-cache enhanced test-time adaptation framework, \textbf{MCP++}. During the process of achieving the entropy minimization objective, MCP++ dynamically refines visual and textual prototypes while reinforcing their mutual alignment. This effectively bridges the modality gap in the feature space, thereby enhancing zero-shot generalization capabilities.

Our main contributions are summarized as follows:
\begin{itemize}
\item We identify a positive correlation between the performance of the caching mechanism and the compactness of cached samples, offering insights to improve adaptive generalization during testing.

\item We propose a multi-cache enhanced prototype-based TTA method, leveraging the entropy, align, and negative caches to achieve more compact intra-class distributions and improve sample prediction accuracy. 

\item We introduce residual fine‑tuning of visual and textual prototypes within our multi‑cache enhancement framework, resulting in more effective alignment between visual and language modalities.
\end{itemize}
\vspace{-2pt}
\section{Related Works}
\label{sec:relatedwork}
\subsection{Vision Language Models}\label{P}
Recent vision-language models (VLMs), such as CLIP \cite{CLIP}, have aligned visual and textual modalities through contrastive learning, demonstrating impressive capabilities in transferring visual knowledge to downstream tasks. These models \cite{Hu2021ScalingUV,Yang2022VisionLanguagePW,Jia2021ScalingUV,CLIP} map images and text into a shared embedding space, enabling efficient cross-modal tasks like image classification \cite{CoOp,CLIP,TipAdapter} and object detection \cite{Du2022LearningTP,Gu2021OpenvocabularyOD}. However, direct transfer often suffers from performance instability due to distribution shifts between the training and target domains \cite{CLIP}. To address this, researchers have proposed prompt learning methods (e.g., CoOp \cite{CoOp}, MaPle \cite{MaPLe}, PromptKD \cite{li2024promptkd}, CPL \cite{Zhang2024ConceptGuidedPL}) that optimize input prompts to enhance model generalization. Adapter-based fine-tuning methods (e.g., CLIP-Adapter \cite{Gao2021CLIPAdapterBV}, Tip-Adapter \cite{TipAdapter}, TaskRes \cite{Yu2022TaskRF}, ClusterAdapter \cite{clusteradapter}, SimNL \cite{Zhang2024EnhancingVF}) further adapt visual or textual representations to specific tasks. Unlike prior work that relies on labeled target‑domain data, we adapt VLMs at test time using only unlabeled samples during inference.

\vspace{-5pt}

\subsection{Prompt-based Test-Time Adaptation}
Due to the cross-modal nature and large-scale parameters of vision-language models, traditional test-time adaptation methods pose significant challenges. Recently, researchers have explored prompt-based strategies tailored for vision-language models \cite{TPT,CoOp,Feng2023DiverseDA,Yoon2024CTPTCT,Sheng2025RTPTIA,Zhang2024HistoricalTP,PromptAlign,meng2025black,wang2025ctpt,zhu2024efficient}. TPT \cite{TPT} was the first to implement test-time adaptation by enforcing consistent predictions across different augmented views, thereby achieving a CoOp \cite{CoOp}-like prompt tuning. DiffTPT \cite{Feng2023DiverseDA} leverages pre-trained diffusion models to enhance the diversity of augmented views. C-TPT \cite{Yoon2024CTPTCT} addresses calibration errors. R‑TPT \cite{Sheng2025RTPTIA} shifts test‑time adaptation from global to per‑sample entropy minimization. HisTPT \cite{Zhang2024HistoricalTP} enhances online TTA by leveraging memory banks with adaptive retrieval, mitigating knowledge forgetting during continuous domain shifts.
Although these prompt-based methods update the model online during inference to better handle distribution shifts, they often come with high computational costs and are time-consuming.

\vspace{-5pt}

\subsection{Cache-enhanced Test-Time Adaptation}
To mitigate the computational expense of prompt-based approaches, an increasing number of cache-based test-time adaptation strategies have been proposed \cite{TDA,Zhang2024BoostAdapterIV,Zhang2024DualMN,DPE,qiao2025bidirectional,zhai2025mitigating}. For instance, TDA \cite{TDA} achieves efficient adaptation by storing both reliable and uncertain test samples. DMN \cite{Zhang2024DualMN} constructs a dynamic memory to capture samples from the test data stream. BoostAdapter \cite{Zhang2024BoostAdapterIV} caches augmented views from a subset of test samples. DPE \cite{DPE} simultaneously computes and evolves multimodal prototypes via caching over time, and is the first to introduce multimodal residual tuning for test-time adaptation.  BPRE \cite{qiao2025bidirectional} further improves cached prototype quality by introducing a multi-dimensional quality-aware reward module and a prototype–reward interactive evolution mechanism. CRG \cite{zhai2025mitigating} mitigates noisy labels in the cache through negative caching and class-wise gaussian modeling.
However, most of these methods rely solely on the entropy of samples to gauge trustworthiness, without fully considering the overall test data distribution.

\section{Method}
\begin{figure*}[h]
  \centering
  \includegraphics[width=\linewidth]{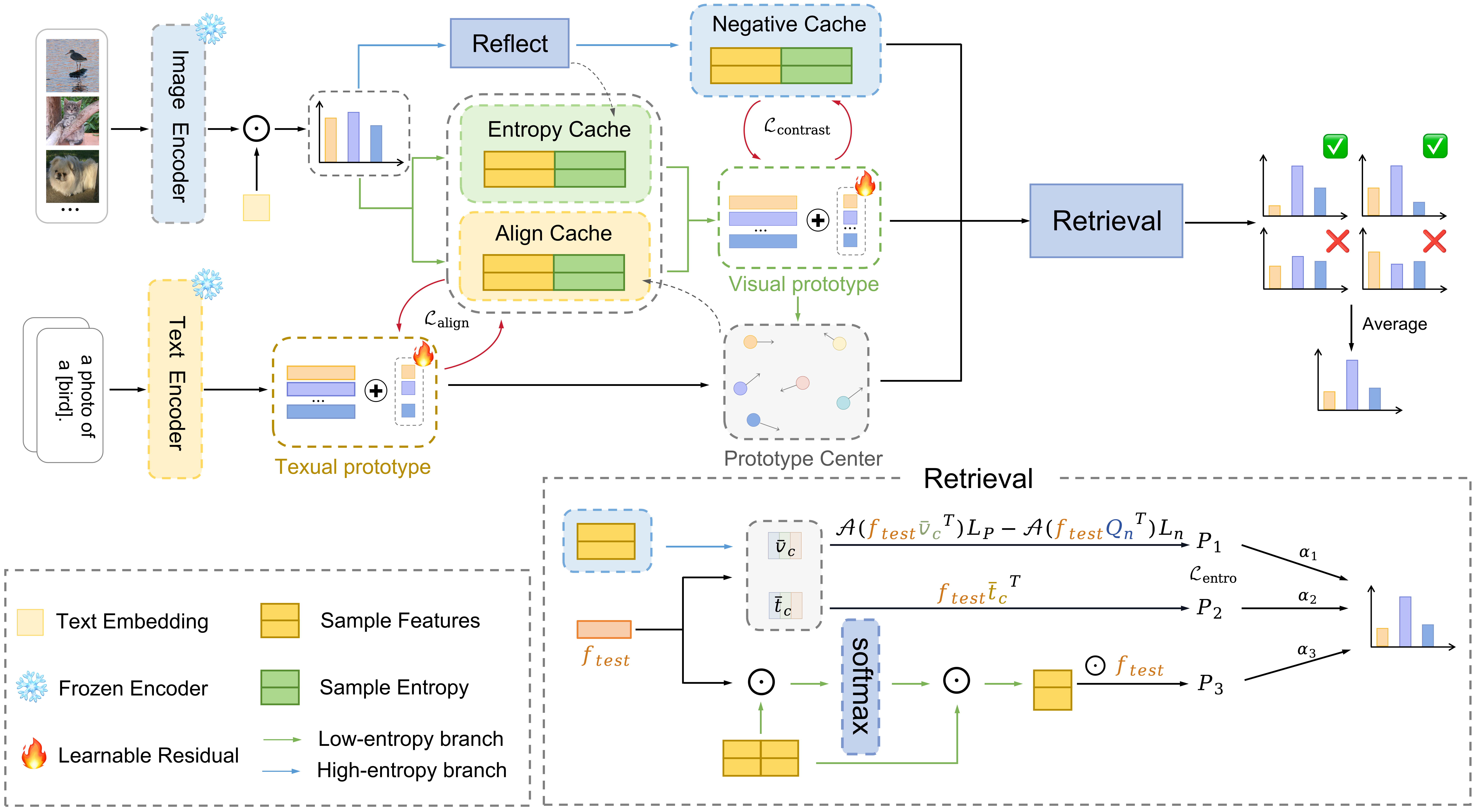} 
  \caption{An overview of our MCP++ method. The Entropy Cache stores low-entropy samples, the Align Cache retains samples closest to the prototype center, and the Negative Cache preserves high-entropy samples after pseudo-label refinement through a reflecting mechanism. We introduce textual and visual prototypes refined with learnable residuals to construct the prototype center, which is optimized using alignment loss, contrastive loss and entropy loss. The final prediction is derived through a retrieval mechanism that aggregates similarity scores from negative cache features, prototype center and adaptive cache features.}
  \vspace{-10pt}
  \label{fig:overview}
\end{figure*}
\subsection{Preliminaries}\label{P}
\textbf{CLIP} \cite{CLIP} is a vision-language model that leverages contrastive learning to align visual and textual representations within a shared embedding space, using a visual encoder $E_v(\cdot)$ and a text encoder $E_t(\cdot)$. Given a test image $x \in X_{\text{test}}$ and its corresponding class-specific textual description $T_c$, CLIP computes the text embedding $t_c = E_t(T_c)$ and the image feature $f_{\text{test}} = E_v(x)$, then determines the class probabilities and the predicted class as follows:
\begin{equation}
\hat{y} = \arg\max_c p_c, \quad p_c = \frac{\exp(\cos(f_{\text{test}}, t_c) / \tau)}{\sum_{j=1}^C \exp(\cos(f_{\text{test}}, t_j) / \tau)},
\end{equation}
where $\cos(\cdot)$ denotes cosine similarity, and $\tau$ is the temperature parameter. 

\subsection{Multi-Cache Enhanced Prototype}\label{C}
Existing cache-based test-time adaptation methods mainly rely on low-entropy samples to construct caches, aiming to stabilize category representations under novel distributions. However, solely relying on low-entropy samples may fail to capture sufficiently compact category representations, particularly for categories with dispersed distributions. To address this issue, we propose the Multi-Cache Enhanced Prototype strategy, as shown in Fig.~\ref{fig:overview}, which dynamically aligns and optimizes visual and textual prototypes through three complementary cache modules, effectively leveraging multimodal information to build more accurate category representations.

Specifically, MCP consists of three modules: the entropy cache stores low-entropy samples to initialize and anchor category representations; the align cache selects samples close to the prototype centers to enhance intra-class compactness; and the negative cache uses negative pseudo-labels to filter high-entropy samples, capturing critical boundary information to reduce noise interference. Next, we will detail the data collection strategies for each cache module and explain how they collaboratively enhance the model's test-time adaptation capability.

\textbf{Entropy Cache.} The entropy cache serves as a dynamically updated low-entropy sample storage mechanism, prioritizing the highest-confidence samples to provide stable category representations and enhance the model’s generalization in dynamic test data \cite{TDA}. For each category, the entropy cache maintains a fixed-capacity queue of size $M (M \ll |X_{\text{test}}|)$, storing only the most representative low-entropy samples. When a new test sample $x$ arrives, if the cache for its predicted category $\hat{y}$ is full, its entropy value H(x) is compared with the highest-entropy sample $x_{\text{max}}$ in the cache, where $H_{\max}^{\hat{y}} = H(x_{\text{max}})$. If $H(x) < H_{\max}^{\hat{y}}$, $x$ replaces $x_{\text{max}}$, ensuring that the cache consistently stores the most reliable low-entropy samples. This mechanism plays a critical role in the test-time adaptation process, preventing prototype shift by anchoring high-confidence features to maintain category stability while facilitating the construction and optimization of category prototypes. However, due to the distribution shifts between the pre-trained model and downstream datasets, relying solely on low-entropy samples may not fully capture the intra-class structural information—especially for categories with dispersed distributions—thereby affecting the accuracy of category prototypes and the model’s adaptability.

\textbf{Align Cache.} To address the limitation of relying solely on low-entropy samples, we construct prototype centers by integrating textual and visual prototypes and introduce an align cache to enhance intra-class compactness. Specifically, for each category $c$, we first compute the textual prototype $\bar{t}_c$ by averaging the embeddings $E_t(\text{p}_i)$ of all textual prompt descriptions $\text{p}_i \in T_c$. The visual prototype $\bar{v}_c$ is obtained by averaging the visual sample features stored in the cache, where the cache sample set, denoted as $M_c$, consists of samples from both the entropy cache $M_c^{\text{entropy}}$ and the align cache $M_c^{\text{align}}$. This formulation allows us to obtain a set of multi-modal prototypes for all categories, expressed as $\mathcal{T} = [\bar{t}_c | c = 1, …, C], \mathcal{V} = [\bar{v}_c | c = 1, …, C] \in \mathbb{R}^{C \times d}$. Subsequently, we form the category prototype center $\mu_c$ by computing a weighted combination of textual and visual prototypes:
\begin{equation}
\mu_c = w\bar{v}_c + (1-w)\bar{t}_c,\label{eq:center}
\end{equation}
where $w \in [0,1]$ balances the contribution between modalities.

When a new test sample $x$ arrives, the align cache first checks whether its entropy satisfies the low-entropy admission criterion, i.e., 
\[
H(x) < H_{\max}^{\hat{y}},
\]
where $H_{\max}^{\hat{y}}$ denotes the highest entropy value among samples stored in the align cache for category $\hat{y}$. If this condition is met, we further calculate the Euclidean distance between the feature embedding of the test sample $f_{\text{test}}$ and the corresponding category prototype center $\mu_{\hat{y}}$, denoted as
$
d(f_{\text{test}}, \mu_{\hat{y}}),
$
and compare it with the distance of the highest-entropy cached sample $f_{\max}$ to the prototype center, $d(f_{\max}, \mu_{\hat{y}})$. Only if
\[
d(f_{\text{test}}, \mu_{\hat{y}}) < d(f_{\max}, \mu_{\hat{y}})
\]
is the cached sample replaced by the new test sample. This replacement strategy ensures that the cache always retains samples that are more reliable and closer to the prototype center, thereby improving the accuracy of category prototypes and enhancing the model’s test-time adaptability.

\textbf{Negative Cache.} The negative cache employs negative pseudo-labels to indicate category absence, thereby mitigating the adverse effects of noisy pseudo-labels \cite{TDA}. However, high-entropy samples inherently exhibit significant uncertainty, and directly using their pseudo-labels may introduce errors. To address this, we introduce a reflecting mechanism prior to storing samples in the negative cache. This mechanism re-evaluates and recalibrates the pseudo-labels of high-entropy samples using features from reliable samples in the existing cache, ensuring that only samples that remain high-entropy after calibration are used for negative learning. This approach more accurately conveys information about category absence, reduces the impact of noise, and enhances the overall adaptability of the model.

Specifically, for each high-entropy sample \(x\), we determine its handling based on its calibrated entropy \(H'(x)\):
\begin{enumerate}
    \item If \(H_{\text{low}} \leq H'(x) \leq H_{\text{high}}\), then sample \(x\) is stored in the negative cache;
    \item If \(H'(x) < H_{\text{low}}\), then sample \(x\) is deemed reliable and may be considered for storage in the entropy cache;
    \item If \(H'(x) > H_{\text{high}}\), then sample \(x\) is discarded.
\end{enumerate}

This criterion is intended to alleviate the risk of prediction errors caused by high entropy or biases in predictions characterized by very low entropy, by integrating test samples that exhibit moderate prediction uncertainty. Ultimately, during inference, we use the features from the negative cache as negative reference signals, which, together with the high-confidence features from the entropy and align caches, participate in the final prediction.

\textbf{Inference.} The model must balance local adaptability with global category stability during inference. To this end, we introduce three complementary information sources in the logits calculation, fully utilizing the prototype center and multi-cache information. Additionally, we design targeted retrieval strategies for different caches to ensure effective feature adaptation and robust classification under varying test-time distributions.

The entropy cache and align cache in the low-entropy pathway store high-confidence representative sample features, which serve as reliable category representations. To dynamically utilize these samples, inspired by the attention mechanism \cite{vaswani2017attention,Zhang2024DualMN,TipAdapter}, we adopt a weighted retrieval mechanism that enables the model to adaptively adjust retrieved category representations based on test samples. Based on similarity calculations, the adaptive cache feature matrix $f_r$ for a specific category $c$ is computed as:

\begin{equation}
    f_r^c = \sum_{f_{c,i} \in M_c} A\big(\cos(f_{\text{test}}, f_{c,i})\big)f_{c,i},
\end{equation}
 where $M_c$ represents the cached samples for category  $c$, including those in the entropy cache and align cache. Each $f_{c,i} \in M_c$ represents a cached feature belonging to class  $c$ , and  $A(x) = \alpha \exp(-\beta (1 - x))$  is the similarity function \cite{TipAdapter}, with  $\alpha$  and  $\beta$  controlling balance and similarity scaling sharpness.

However, relying solely on cached information may introduce local biases in category representation, affecting class discrimination. To address this, we introduce textual and visual prototypes from the prototype center to provide comprehensive multi-modal information, while utilizing the negative cache feature matrix $Q_n$ to suppress uncertain categories and mitigate misclassification risks. During the similarity computation process, we first determine the matching degree between the visual prototype and the test sample, and then refine the category prediction by incorporating information from the negative cache:
\begin{equation}
    P(f_{\text{test}}, \mathcal{V}, Q_n) = A(f_{\text{test}} \mathcal{V}^\top) L_p - A(f_{\text{test}} Q_n^\top) L_n,
\end{equation}
here, $L_p$ represents the one-hot vector of the predicted class, while $L_n$ denotes the negative class probability mask defined in the TDA \cite{TDA}.
Finally, the prediction logits are obtained by a weighted combination of three normalized complementary components:
\begin{equation}
 p(f_{\text{test}}) = \alpha_1 f_{\text{test}} \mathcal{T}^\top + \alpha_2 P(f_{\text{test}}, \mathcal{V}, Q_n) + \alpha_3 f_{\text{test}} f_r^\top ,\label{eq:logit}
\end{equation}
where  $\alpha_1, \alpha_2, \alpha_3$  are weights controlling contributions from textual semantic matching, visual-category contrastive information, and cache retrieval features. This retrieval strategy effectively integrates multi-source information, enhancing class discrimination and feature stability, thereby achieving robust generalization across different test distributions.

\subsection{MCP++: Residual-Enhanced Multimodal Prototype Optimization Mechanism}\label{CC}

\begin{table*}[h]
    \centering
    \renewcommand{\arraystretch}{1.2}  
    \setlength{\tabcolsep}{5.8pt}        
    \fontsize{9}{8.5}\selectfont
    \caption{Results on the Cross-Domain Benchmark. Top-1 accuracy (\%) results are presented for all evaluated methods employing the ViT-B/16 visual backbone of CLIP. The best results are highlighted in bold.}
    \label{tab:cross_domain}
    \begin{tabular}{l|*{10}{c}|c}
    \toprule
    \multicolumn{1}{l}{\textbf{Method}} & 
    \multicolumn{1}{c}{\textbf{Aircraft}} & 
    \multicolumn{1}{c}{\textbf{Caltech}} & 
    \multicolumn{1}{c}{\textbf{Cars}} & 
    \multicolumn{1}{c}{\textbf{DTD}} & 
    \multicolumn{1}{c}{\textbf{EuroSAT}} & 
    \multicolumn{1}{c}{\textbf{Flower}} & 
    \multicolumn{1}{c}{\textbf{Food101}} & 
    \multicolumn{1}{c}{\textbf{Pets}} & 
    \multicolumn{1}{c}{\textbf{SUN397}} & 
    \multicolumn{1}{c}{\textbf{UCF101}} & 
    \multicolumn{1}{c}{\textbf{Average}} \\
    \midrule
    CLIP-ViT-B/16 & 23.67 & 93.35 & 65.48 & 44.27 & 42.01 & 67.44 & 83.65 & 88.25 & 62.59 & 65.13 & 63.58 \\
    \rowcolor{mygray} CoOp & 18.47 & 93.70 & 64.51 & 41.92 & 46.39 & 68.71 & 85.30 & 89.14 & 64.15 & 66.55 & 63.88 \\
    \rowcolor{mygray} MaPle & 24.74 & 93.53 & 65.57 & 46.49 & 48.06 & 72.23 & 86.20 & 90.49 & 67.01 & 68.69 & 66.30 \\
    \midrule
    \rowcolor{mypink2} TPT & 24.78 & 94.16 & 66.87 & 47.75 & 42.44 & 68.98 & 84.67 & 87.79 & 65.50 & 68.04 & 65.10 \\
    \rowcolor{mypink2} DiffTPT & 25.60 & 92.49 & 67.01 & 47.00 & 43.13 & 70.10 & \textbf{87.23} & 88.22 & 65.74 & 62.67 & 65.47 \\
    \rowcolor{mypink2} PromptAlign & 24.80 & 94.01 & 68.50 & 47.24 & 47.86 & 72.39 & 86.65 & 90.76 & 67.54 & 69.47 & 66.92 \\
    \midrule
    \rowcolor{mypink1} TDA & 23.91 & 94.24 & 67.28 & 47.40 & 58.00 & 71.42 & 86.14 & 88.63 & 67.62 & 70.66 & 67.53 \\
    \rowcolor{mypink1} DPE & 28.95 & 94.81 & 67.31 & 54.20 & 55.79 & 75.07 & 86.17 & 91.14 & 70.07 & 70.44 & 69.40 \\
    \rowcolor{mypink1} BoostAdapter & 27.45 & 94.77 & 69.30 & 45.69 & 61.22 & 71.66 & 87.17 & 89.51 & 68.09 & 71.93 & 68.68 \\
    \rowcolor{mypink1} DMN & 30.03 & 95.38 & 67.96 & 55.85 & 59.43 & 74.49 & 85.08 & 92.04 & 70.18 & 72.51 & 70.30\\
    \midrule
   \rowcolor{mypink} \textbf{MCP}   & 30.18 & 95.33 & 69.99 & 55.91 & 68.42 & 75.88 & 86.85 & 91.88 & 71.04 & 74.36 & 71.98\\
    \rowcolor{mypink} \textbf{MCP++} & \textbf{31.06} & \textbf{95.50} & \textbf{70.13} & \textbf{56.97} & \textbf{68.69} & \textbf{77.55} & 87.20 & \textbf{92.40} & \textbf{71.17} & \textbf{75.44} & \textbf{72.61}\\
    \bottomrule
    \end{tabular}
    
\end{table*}
Although MCP has improved test-time adaptation and stable performance, its constructed prototypes still suffer from insufficient cross-modal alignment. To address this issue, we propose MCP++, which incorporates a vision-text prototype residual learning mechanism into the multi-cache system to align and dynamically optimize the prototypes. Thus, it effectively bridges the modality gap in the feature space and enhances the generalization ability in zero-shot classification.

Specifically, we introduce learnable residual parameters to refine the visual and textual prototypes. These residuals are directly added to their respective prototypes. After each test sample $x$ is incorporated into and updates the cache, we compute two sets of prototypes $\mathcal{T} = [\bar{t}_c \mid c = 1, \dots, C]$ and $\mathcal{V} = [\bar{v}_c \mid c = 1, \dots, C]$. Then, the residual parameters initialized to zero are dynamically optimized through gradient updates. The final refined prototypes are expressed as:
\begin{equation}
    \bar{t}_c' = \bar{t}_c + R_t^c,\quad \bar{v}_c' = \bar{v}_c + R_v^c
\end{equation}
where $R_t, R_v \in \mathbb{R}^{C \times d}$ are learnable residual parameters applied to the textual and visual prototypes, respectively, to further optimize category representations.

To jointly optimize the residual parameters and prototype centers, we design a weighted loss function integrating multiple objectives. Inspired by the TPT \cite{TPT}, we first introduce an unsupervised entropy minimization loss $\mathcal{L}_{\text{entro}}$, which reduces prediction uncertainty by encouraging consistent predictions across different high-confidence augmented views of each test sample. The loss is defined as follows:
\begin{equation}
\mathcal{L}_{\text{entro}} = H\left( \frac{1}{\rho N} \sum_{n=1}^N \mathds{1}\Big[ H\big(P_n\big) \leq \delta \Big] P_n\right),
\label{eq:entro}
\end{equation}
where $H(\cdot)$ denotes the entropy function, $N$ is the number of augmented views per test sample, $P_n=P(A_n(x))$ represents the prediction of the $n$-th augmented images, $\delta$ is the low-entropy threshold, and $\rho$ denotes the proportion of selected high-confidence samples.

Although entropy minimization helps reduce prediction uncertainty, it does not address the issue of feature alignment across modalities. To this end, motivated by \cite{DPE,clusteradapter}, we incorporate a vision-text alignment loss $\mathcal{L}_{\text{align}}$ to align visual prototypes with their corresponding textual embeddings, formulated based on the InfoNCE~\cite{infoNCE} loss as follows:
\begin{equation}
    \begin{split}
        &\mathcal{L}_{\text{align}} = \frac{1}{C} \sum_{c=1}^{C} S_c, \text{where} \\
        &S_c = -\log \frac{\exp (\bar{t}_{c}'^\top \bar{v}_{c}')}{\sum\limits_{j=1}^{C} \exp (\bar{t}_{c}'^\top \bar{v}_{j}')}  -\log \frac{\exp (\bar{t}_{c}'^\top \bar{v}_{c}')}{\sum\limits_{j=1}^{C} \exp (\bar{t}_{j}'^\top \bar{v}_{c}')},
    \end{split}
\end{equation}
where $C$ denotes the total number of categories.

The negative cache stores relatively high-entropy samples, which typically reside near category boundaries and exhibit high uncertainty. To effectively leverage these samples, we introduce a positive-negative contrastive loss $\mathcal{L}_{\text{contrast}}$, which increases the distance between prototype centers and high-entropy negative samples. This prevents them from interfering with category representations, reduces inter-class confusion, and enhances the model’s generalization capability at test time. The loss is defined as:
\begin{equation}
\mathcal{L}_{\text{contrast}} = -\log \left( 1 - \frac{1}{C} \sum_{c=1}^C \cos\left(\bar{v}_c', \bar{v}^{neg}_{c}\right) + \epsilon \right),
\end{equation}
where $\bar{v}^{neg}_{c}$ represents the mean feature embedding of class $c$ in the negative cache, and $\epsilon$ is a small constant for numerical stability.

The final joint loss function integrates the three objectives:
\begin{equation}
\mathcal{L} = \mathcal{L}_{\text{entro}} + \lambda \cdot \mathcal{L}_{\text{align}} + \gamma \cdot \mathcal{L}_{\text{contrast}},\label{eq:loss}
\end{equation}
where $\lambda$ and $\gamma$ are weighting factors controlling the contributions of each loss term. This unified loss ensures the residuals and prototype centers are jointly optimized, allowing the framework to dynamically adapt to distribution shifts during test time and achieve robust zero-shot classification.

\begin{table*}[t]
    \renewcommand{\arraystretch}{1.2}  
    \setlength{\tabcolsep}{8.5pt}       
    \fontsize{9}{8.5}\selectfont
    \centering
    \caption{Results on the OOD Benchmark. Top-1 accuracy (\%) results are presented for all evaluated methods employing the ViT-B/16 visual backbone of CLIP. The best results are highlighted in bold.}
    \label{tab:ood}
    \begin{tabular}{l|c c c c c|c|c}
    \toprule
    \multicolumn{1}{l}{\textbf{Method}} & 
    \multicolumn{1}{c}{\textbf{ImageNet}} & 
    \multicolumn{1}{c}{\textbf{ImageNet-A}} & 
    \multicolumn{1}{c}{\textbf{ImageNet-V2}} & 
    \multicolumn{1}{c}{\textbf{ImageNet-R}} & 
    \multicolumn{1}{c}{\textbf{ImageNet-S}} & 
    \multicolumn{1}{c}{\textbf{OOD Average}} & 
    \multicolumn{1}{c}{\textbf{Average}} \\
    \midrule
    CLIP-ViT-B/16 & 66.73 & 47.87 & 60.86 & 73.98 & 46.09 & 57.20 & 59.11 \\
    \rowcolor{mygray} CoOp & 71.51 & 49.71 & 64.20 & 75.21 & 47.99 & 59.28 & 61.72 \\
    \rowcolor{mygray} MaPle & 70.72 & 50.90 & 64.07 & 76.98 & 49.15 & 60.69 &  62.36 \\
    \midrule
    \rowcolor{mypink2} TPT & 68.98 & 54.77 & 63.45 & 77.06 & 47.94 & 60.81 & 62.44 \\
    \rowcolor{mypink2} DiffTPT & 70.30 & 55.68 & 65.10 & 75.00 & 46.80 & 60.52 & 62.28 \\
    \midrule
    \rowcolor{mypink1} TDA & 69.51 & \textbf{60.11} & 64.67 & 80.24 & 50.53 & 63.89 & 65.01 \\
    \rowcolor{mypink1} DMN & 72.25 & 58.28 & 65.17 & 78.55 & 53.20 & 63.80 & 65.49  \\
    \rowcolor{mypink1} DPE & 71.91 & 59.63 & 65.44 & 80.40 & 52.26 & 64.43 & 65.93 \\
    \midrule
    \rowcolor{mypink} \textbf{MCP} & 72.37 & 59.71 & 65.16 & 81.59 & 53.89 & 65.09 & 66.54\\
   \rowcolor{mypink} \textbf{MCP++} & \textbf{72.64} & 59.78 & \textbf{65.77} & \textbf{81.73} & \textbf{54.39} & \textbf{65.42} & \textbf{66.86}\\
    \bottomrule
    \end{tabular}
\end{table*}

\vspace{-3pt}
\section{Experiment}

\subsection{Datasets}
To comprehensively evaluate the effectiveness of our method, we conduct experiments on both natural distribution shift and cross-dataset generalization, following the evaluation protocol in TPT \cite{TPT}, and report the top-1 accuracy. For cross-dataset generalization, we evaluate the model’s ability to adapt to different visual domains using 10 diverse datasets: FGVC-Aircraft \cite{aircraft}, Caltech101 \cite{caltech101}, Stanford Cars \cite{cars}, DTD \cite{Dtd}, EuroSAT \cite{helber2019eurosat}, Flowers102 \cite{oxfordflower}, Food101 \cite{bossard2014food}, Oxford Pets \cite{oxford_pets}, SUN397 \cite{sun397}, and UCF101 \cite{ucf101}. To evaluate robustness under natural distribution shifts, we test the model on ImageNet \cite{deng2009imagenet} and its four variants: ImageNet-V2 \cite{recht2019imagenet}, ImageNet-Sketch \cite{wang2019learning}, ImageNet-A \cite{hendrycks2021natural}, and ImageNet-R \cite{hendrycks2021many}. 
\subsection{Implementation details}
We conduct all experiments using ResNet50 and ViT-B/16 as the vision encoders of CLIP \cite{TPT}, on a single 48 GB NVIDIA A40 GPU. To enhance robustness, we generate 31 augmented views per test image through random resized cropping, resulting in a total of 32 images per sample. For optimization, we employ the AdamW optimizer to update the learnable parameters for one step, using a learning rate 0.0001. The cache sizes are uniformly set to $\lvert M_c^{\text{entropy}} \rvert=10$, $\lvert M_c^{\text{align}} \rvert=10$, and $\lvert M_c^{\text{negative}} \rvert=3$. In Eq.~\eqref{eq:center}, the parameter $w$ is set to 0.8, while in Eq.~\eqref{eq:loss}, we set $\lambda=0.5$ and $\gamma=0.2$. For each downstream task, we search the optimal combination of prediction weights $\alpha_1 \sim \alpha_3$ in Eq.~\eqref{eq:logit} following \cite{Zhang2024DualMN}. The proportion of selected high-confidence samples $\delta$ in Eq.~\eqref{eq:entro} is set to 0.1. More experimental results are provided in the Appendix.

\subsection{Comparison to State-of-the-Art}

\textbf{Cross-Dataset Generalization.} Table~\ref{tab:cross_domain} presents the results of our Cross-Dataset Generalization experiments, comparing our method against state-of-the-art approaches across ten fine-grained and specialized datasets. The results demonstrate that our method achieves superior robustness under significant distributional shifts. On the ViT-B/16 backbone, our method achieves the best performance on 9 out of 10 tasks. Compared to prompt-based methods, including CoOp \cite{CoOp}, TPT \cite{TPT}, DiffTPT \cite{Feng2023DiverseDA}, Maple \cite{MaPLe}, and PromptAlign \cite{PromptAlign}, our approach not only achieves substantial improvements but also reduces the computational overhead of optimization. Compared to cache-based methods, both the training-free MCP and the residual learning-based MCP++ achieve superior results. Specifically, MCP++ outperforms DMN-ZS \cite{Zhang2024DualMN}, DPE \cite{DPE} and BoostAdapter \cite{Zhang2024BoostAdapterIV} by 2.31\%, 3.21\% and 3.93\%, respectively. Notably, MCP and MCP++ demonstrate improvements of 4.45\% and 5.08\% over TDA, indicating that incorporating a multi-cache mechanism, which captures a comprehensive feature distribution, can further improve model generalization.

\textbf{Natural Distribution Shifts.} We further evaluate the robustness of our proposed method on in-domain ImageNet and its four Out-Of-Distribution(OOD) variants, with the results presented in Table ~\ref{tab:ood}. Our approach outperforms the leading prompt-tuning methods, achieving a 4.18\% improvement over MaPLe with MCP and a 4.5\% improvement with MCP++. When compared to cache-based methods, MCP++ surpasses TDA, DMN-ZS, and DPE by 1.85\%, 1.37\%, and 0.93\%, respectively, further demonstrating the superiority of our approach. These results indicate that our method is generally effective in both domain-specific variations and out-of-distribution robustness scenarios.

\begin{figure*}[ht]
  \centering
  \begin{subfigure}[b]{0.3\linewidth}
    \includegraphics[width=\linewidth,height=\linewidth]{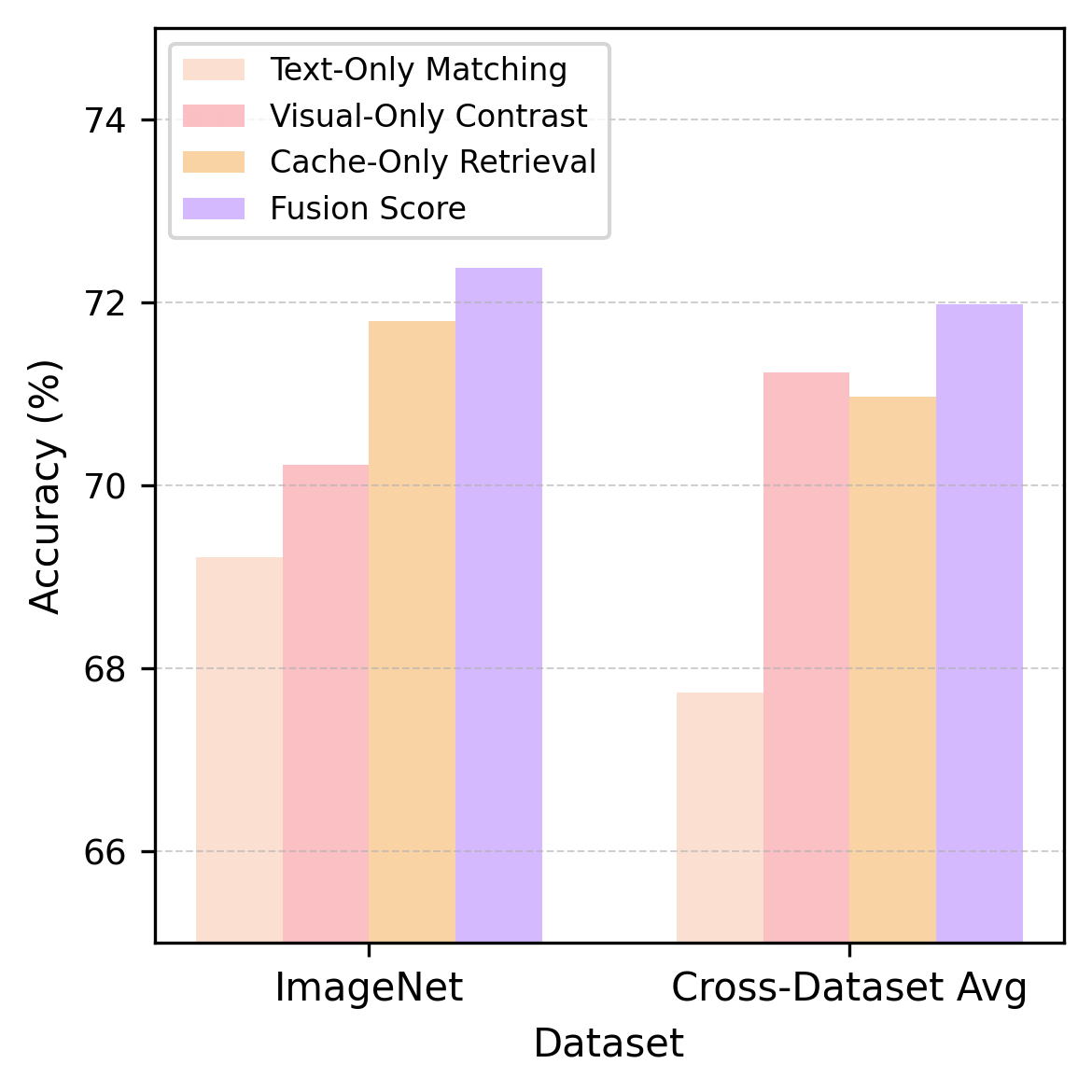}
    \label{fig:sub-a}
  \end{subfigure}
  \hfill
  \begin{subfigure}[b]{0.3\linewidth}
    \includegraphics[width=\linewidth,height=\linewidth]{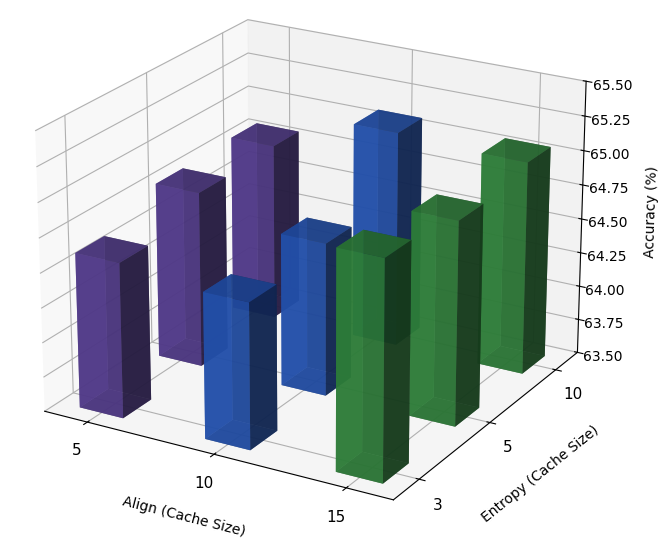}
    \label{fig:sub-b}
  \end{subfigure}
  \hfill
  \begin{subfigure}[b]{0.3\linewidth}
    \includegraphics[width=\linewidth,height=\linewidth]{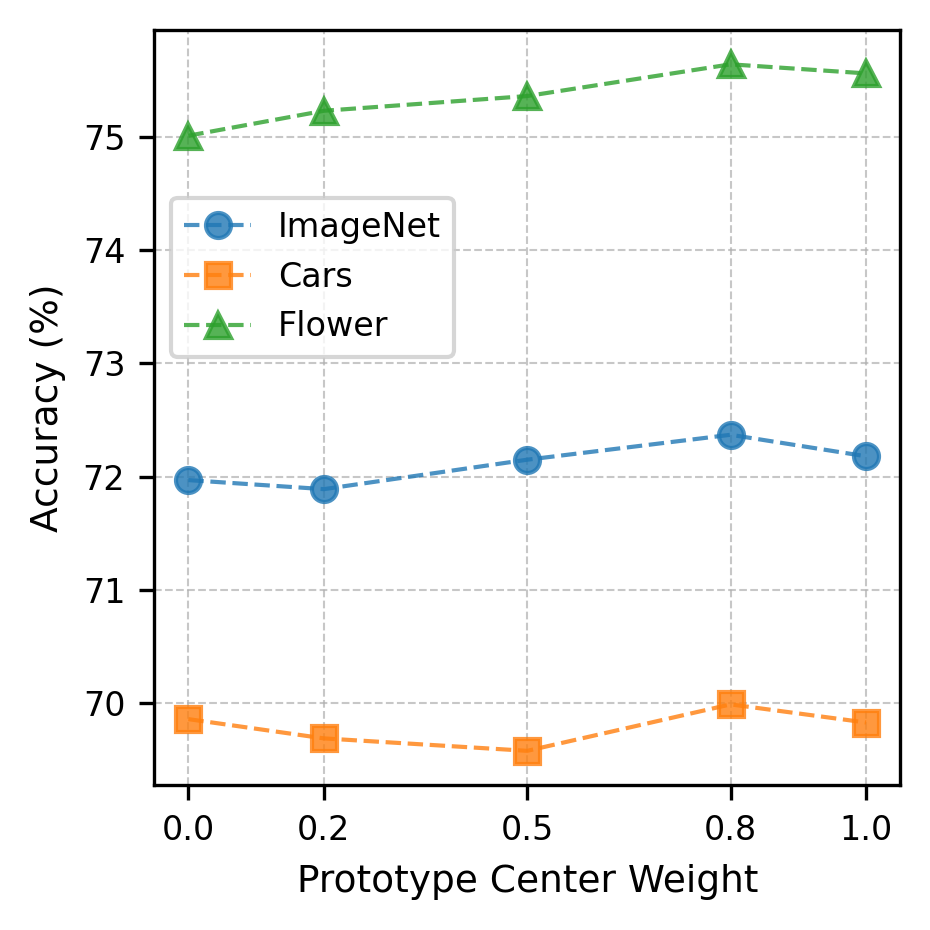}
    \label{fig:sub-c}
  \end{subfigure}
  \vspace{-5pt}
  \caption{Ablation studies. (Left) Effect of retrieval strategies on ImageNet and the average performance across 10 cross-domain datasets. (Middle) Impact of cache size settings on cross-domain datasets. (Right) Performance variation across three datasets with different prototype center weights $w$ in Eq.~\eqref{eq:center}.}
  \label{fig:ablation}
  \vspace{-5pt}
\end{figure*}

\subsection{Ablation Studies}
\textbf{Effectiveness of Multi-Cache Mechanism.}
We conducted ablation experiments on these three cache components and found that each cache functions effectively on its own, significantly outperforming the baseline CLIP model. As shown in Table ~\ref{tab:cacheab}, this result demonstrates that all stored feature representations contribute meaningfully to test-time adaptation. Furthermore, the model achieves the highest performance when all three caches are integrated, highlighting their complementary nature. 

\begin{table}[!htbp]
  \centering
  \caption{Ablation study on the effectiveness of different caches designed in MCP++, evaluated on ImageNet.}
  \begin{tabular}{ccc|c}
    \toprule
    \textbf{Entropy} & \textbf{Align} & \textbf{Negative} & \textbf{ImageNet Acc.} \\
    \midrule
    \checkmark & \checkmark &            & 72.49 \\
    \checkmark &            & \checkmark & 71.73 \\
               & \checkmark & \checkmark & 72.54 \\
               & \checkmark &            & 71.51 \\
    \checkmark &            &            & 72.47 \\
               &            & \checkmark & 71.71 \\
    \rowcolor{gray!15}
    \checkmark & \checkmark & \checkmark & \textbf{72.64} \\
    \bottomrule
  \end{tabular}
  \vspace{-10pt}
  \label{tab:cacheab}
\end{table}


  

\textbf{Impact of Retrieval Mechanism.} To evaluate the robustness of different retrieval strategies, we conduct an ablation study on the three retrieval mechanisms in Eq.~\eqref{eq:logit}, including Text-Only Matching, Visual Prototype Contrast, and Cache-Only Retrieval. As shown in Fig.\ref{fig:ablation} (Left), incorporating visual category information significantly enhances generalization compared to text-only matching, while cache-based retrieval further improves model performance. Ultimately, our multi-retrieval strategy achieves a 3.16\% and 4.24\% improvement on ImageNet and ten cross-domain datasets, respectively, outperforming all single retrieval methods and confirming the complementarity of different information sources. 

\textbf{Effect of Cache Size.} We further conduct a parameter study on the entropy cache and align cache sizes to determine the optimal trade-off between sample diversity and matching accuracy. As shown in Fig.\ref{fig:ablation} (Middle), model performance tends to decline when the cache size is too small or too large, indicating the importance of a balanced configuration. Experimental results show that under the ResNet-50 backbone, the cross-domain datasets achieve the highest average accuracy when $\lvert M_c^{\text{align}} \rvert=10$ and $\lvert M_c^{\text{entropy}} \rvert=10$. A smaller entropy cache limits the number of high-confidence samples, resulting in inaccurate category centers and weaker generalization, while an oversized cache introduces more uncertain samples, compromising distribution reliability. Similarly, a small align cache restricts effective category alignment, whereas an excessively large one increases redundancy, destabilizing decision-making.

\textbf{Modal Feature Fusion in Prototype Centers.} In Eq.~\eqref{eq:center}, we apply a weighted fusion of visual prototypes and textual prototypes to construct a more robust prototype center. To assess the impact of different modal features on classification performance, we conduct an ablation study by adjusting the weight parameter. As shown in Fig.\ref{fig:ablation} (Right), the trend reflects that integrating both modalities contributes to robust performance, indicating the benefit of leveraging complementary visual and textual features. However, we observe a trade-off between the two modalities, where excessively high or low weights negatively affect model adaptability. Through empirical evaluation, we find that when $w=0.8$, the model achieves optimal performance across multiple datasets.

\vspace{-5pt}
\section{Conclusion}
In this study, we first identify a positive correlation between cache-based test-time adaptation performance and intra-class compactness, demonstrating that a more compact intra-class distribution effectively enhances generalization. To leverage this insight, we propose MCP, which refines category prototype construction through the entropy cache, align cache, and negative cache. Building on this foundation, MCP++ further integrates cross-modal prototype alignment and residual learning, optimizing the fusion of visual and textual features to enhance model adaptability. Experimental results show that our approach outperforms existing methods across 15 downstream tasks, validating the critical role of considering intra-class compactness and capturing the comprehensive feature distribution of test data in cache-based adaptation, while also demonstrating the effectiveness of the multi-cache mechanism. This work advances research and applications in test-time adaptation and provides a promising solution for improving the generalization of vision-language models under distribution shifts.
{
    \small
    \bibliographystyle{ieeenat_fullname}
    \bibliography{main}
}
\clearpage                      
\appendix                       
\renewcommand\thesection{\Alph{section}} 
\twocolumn[{
  \centering
  {\Large\bfseries
    Multi-Cache Enhanced Prototype Learning for Test-Time Generalization of Vision-Language Models\par
  }
  \vspace{0.5em}
  {\large Supplementary Material\par}
  \vspace{1.5em}     
}]

\appendix

In appendix, we provide additional details and experimental results to enhance understanding and insights into our proposed method.
This supplementary document is organized as follows:
\begin{itemize}[leftmargin=0.5cm, itemindent=0cm, itemsep=4pt,topsep=4pt,parsep=0pt]
    \item[$\bullet$] \textbf{Detailed Dataset Information:}  
    Comprehensive details about the datasets used in our experiments, including their key characteristics and distributions, are provided.
    \item[$\bullet$]  \textbf{Text Templates for Each Dataset:}  
    The text templates used in our experiments for each dataset are listed for reproducibility.
    \item[$\bullet$]  \textbf{Further Discussion on Related Work:}  
    We presented further details of the baselines and highlighted the distinctions between our approach and theirs.
    \item[$\bullet$]  \textbf{Additional Experimental Results:}
    We provide the accuracy of our method based on the ResNet-50 backbone on natural distribution shifts and cross-dataset generalization, and also provide ablation experimental results of alignment loss, contrast loss, and sensitivity analysis of negative cache size.
     \item[$\bullet$] \textbf{Theoretical Analysis:} A formal derivation is provided to demonstrate that, compared to using an entropy cache alone, the addition of an align cache results in a lower excess error bound, thereby improving generalization performance.
    
\end{itemize}
\section{Detailed Dataset Information}
In Table \ref{tab:dataset}, we provide comprehensive statistics for each dataset utilized in our experiments, detailing the number of classes, the sizes of the training, validation, and test sets, as well as their associated original tasks. These datasets have emerged as key benchmarks for evaluating the test-time adaptation of vision-language models \cite{TPT,TPS,TDA,Feng2023DiverseDA}.
\begin{table}[ht]
  \caption{Statistics of datasets used in our experiments.}
  \label{tab:dataset}
  \centering
  \resizebox{\columnwidth}{!}{%
    \setlength{\tabcolsep}{1.5mm}%
    \begin{tabular}{lccccc}
      \toprule
      Dataset & Classes & Train & Val & Test & Task \\
      \midrule
      \multicolumn{6}{l}{\emph{Standard benchmarks}} \\[2pt]
      Caltech101~\cite{caltech101}        & 100   & 4\,128  & 1\,649 & 2\,465  & Object recognition \\
      DTD~\cite{Dtd}                      & 47    & 2\,820  & 1\,128 & 1\,692  & Texture recognition \\
      EuroSAT~\cite{helber2019eurosat}    & 10    & 13\,500 & 5\,400 & 8\,100  & Satellite scene classification \\
      FGVCAircraft~\cite{aircraft}        & 100   & 3\,334  & 3\,333 & 3\,333  & Aircraft type recognition \\
      Flowers102~\cite{oxfordflower}      & 102   & 4\,093  & 1\,633 & 2\,463  & Flower species classification \\
      Food101~\cite{bossard2014food}      & 101   & 50\,500 & 20\,200& 30\,300 & Food image classification \\
      ImageNet~\cite{deng2009imagenet}    & 1\,000& 1.28\,M & —     & 50\,000 & Large‑scale object recognition \\
      OxfordPets~\cite{oxford_pets}       & 37    & 2\,944  & 736   & 3\,669  & Pet breed recognition \\
      StanfordCars~\cite{cars}            & 196   & 6\,509  & 1\,635 & 8\,041  & Car model recognition \\
      SUN397~\cite{sun397}                & 397   & 15\,880 & 3\,970 & 19\,850 & Scene recognition \\
      UCF101~\cite{ucf101}                & 101   & 7\,639  & 1\,898 & 3\,783  & Action recognition \\
      \midrule
      \multicolumn{6}{l}{\emph{Robustness benchmarks}} \\[2pt]
      ImageNet‑V2~\cite{recht2019imagenet}& 1\,000 & — & — & 10\,000  & Collocation robustness \\
      ImageNet‑A~\cite{hendrycks2021natural}& 200  & — & — & 7\,500   & Natural adversarial robustness \\
      ImageNet‑R~\cite{imagenet-r}        & 200   & — & — & 30\,000  & Multi‑domain robustness \\
      ImageNet‑S~\cite{imagenet-s}        & 1\,000 & — & — & 50\,899  & Sketch‑domain robustness \\
      \bottomrule
    \end{tabular}%
  }

\end{table}
\section{Text Templates for Each Dataset}
In Table~\ref{tab:prompt}, we detail the specific hand-crafted prompts utilized for each dataset, following previous work \cite{TDA}. We also employ CuPL \cite{Pratt2022WhatDA} prompts to further enhance performance.

\begin{table}[t]
\centering
    \caption{\looseness=-1 {Textual prompts template used in experiments}.}
    \label{tab:prompt}
    \resizebox{\columnwidth}{!}{%
    \setlength{\tabcolsep}{8mm}%
    \begin{tabular}{lc}
    \toprule
    Dataset                  & Prompts   \\ \midrule
    Caltech101~\cite{caltech101} & ``a photo of a \{\texttt{CLASS}\}.'' \\
    DTD~\cite{Dtd} & ``\{\texttt{CLASS}\} texture.'' \\ 
    EuroSAT~\cite{helber2019eurosat} & ``a centered satellite photo of \{\texttt{CLASS}\}.'' \\ 
    FGVCAircraft~\cite{aircraft} & ``a photo of a \{\texttt{CLASS}\}, a type of aircraft.'' \\
    Flowers102~\cite{oxfordflower} & ``a photo of a \{\texttt{CLASS}\}, a type of flower.'' \\ 
    Food101~\cite{bossard2014food} & ``a photo of \{\texttt{CLASS}\}, a type of food.'' \\ 
    OxfordPets~\cite{oxford_pets} & ``a photo of a \{\texttt{CLASS}\}, a type of pet.''  \\ 
    StanfordCars~\cite{cars} & ``a photo of a \{\texttt{CLASS}\}.'' \\
    SUN397~\cite{sun397} & ``a photo of a \{\texttt{CLASS}\}.''\\ 
    UCF101~\cite{ucf101} & ``a photo of a person doing \{\texttt{CLASS}\}.'' \\
    \midrule
    & ``itap of a \{\texttt{CLASS}\}.'' \\ 
    ImageNet~\cite{deng2009imagenet} & ``a origami \{\texttt{CLASS}\}.'' \\ 
    ImageNet-S~\cite{imagenet-s} & ``a bad photo of the \{\texttt{CLASS}\}.'' \\ 
    ImageNet-R~\cite{imagenet-r} &  ``a photo of the small \{\texttt{CLASS}\}.''\\ 
    ImageNet-V2~\cite{recht2019imagenet} & ``a photo of the large \{\texttt{CLASS}\}.'' \\ 
    ImageNet-A~\cite{hendrycks2021natural} & ``a \{\texttt{CLASS}\} in a video game.'' \\ 
    & ``art of the \{\texttt{CLASS}\}.'' \\  
    \bottomrule
    \end{tabular}%
    }
\end{table}

\section{Further Discussion on Related Work}

\begin{table*}[h]
    \centering
    \renewcommand{\arraystretch}{1.2}  
    \setlength{\tabcolsep}{8pt}        
    \fontsize{8.5}{8.5}\selectfont
    \caption{Comparison of different methods. \textbf{Multimodal}: whether multimodal learning is employed; \textbf{HistInfo}: whether historical information is utilized; \textbf{NoEncGrad}: no access to encoder gradients; \textbf{NoExtraData}: no requirement for extra data or models;  \textbf{HighEntro}: leverages information from high-entropy samples ; \textbf{ExpTestDist}: explicitly considers the distribution of test samples.}
    \label{tab:comparison}
    \begin{tabular}{lcccccc}
    \toprule
    Method & Multimodal & HistInfo & NoEncGrad & NoExtraData & HighEntro & ExpTestDist  \\
    \midrule
    \rowcolor{mypink2} TPT & \xmark & \xmark & \xmark & \cmark & \xmark & \xmark  \\
    \rowcolor{mypink2} DiffTPT & \xmark & \xmark & \xmark & \xmark & \xmark & \xmark  \\
    \rowcolor{mypink2} PrompAlign & \cmark & \xmark & \xmark & \xmark & \xmark & \cmark  \\
    \midrule
    \rowcolor{mypink1} TDA & \xmark & \cmark & \cmark & \cmark & \cmark & \xmark   \\
    \rowcolor{mypink1} DMN & \xmark & \cmark & \cmark & \cmark & \xmark & \xmark  \\
    \rowcolor{mypink1} BoostAdapter & \xmark & \cmark & \cmark & \cmark & \xmark & \xmark  \\
    \rowcolor{mypink1} DPE & \cmark & \cmark & \cmark & \cmark & \xmark & \xmark  \\
    \midrule
    \rowcolor{mypink} \textbf{MCP} & \xmark & \cmark & \cmark & \cmark & \cmark & \cmark \\
    \rowcolor{mypink} \textbf{MCP++} & \cmark & \cmark & \cmark & \cmark & \cmark & \cmark \\
    \bottomrule
    \end{tabular}
\end{table*}

We acknowledge that our approach shares some high-level similarities with certain existing methods; however, there are some critical differences. In the following, we discuss the distinctions between our method TDA \cite{TDA}, DPE \cite{DPE} and PromptAlign \cite{PromptAlign} individually, and further compare various approaches from several perspectives as summarized in \ref{tab:comparison}. Moreover, following these comparisons, we devote an entire chapter to Prototype Learning to further discuss prototype‑based research.

\subsection{TDA \cite{TDA}}
While both DMN \cite{Zhang2024DualMN} and TDA \cite{TDA} enhance test-time generalization by storing historical test samples in a cache or memory module, they share some similarities in this regard. However, these methods rely solely on entropy to assess sample quality, which limits the potential benefits of caching. In contrast, our MCP method fully leverages category distribution information by introducing an alignment cache to promote intra-class compactness, thereby further enhancing the performance gains from caching. Moreover, MCP++ incorporates learnable residual parameters to refine prototype representations, enabling the model to dynamically adapt based on test samples.
\subsection{DPE \cite{DPE}}
Similarly, DPE \cite{DPE} relies solely on entropy as the criterion for caching samples and does not utilize high-entropy samples within the test stream. Its performance gains primarily come from evolving visual and textual prototypes to enhance the model’s generalization capability during testing. In contrast, our MCP method requires no additional training and achieves superior performance and higher efficiency than DPE by fully exploiting the complementary multi-cache mechanism. 

Similar to DPE, MCP++ augments both visual and textual prototypes with learnable residual vectors, and performs multimodal fine-tuning. However, the three‑way cache (Entropy, Align, Negative) supplies a diverse, complementary test sample set, and the \textbf{Prototype Center} constraint pushes the visual prototypes to high quality before any multimodal residual tuning. As a result, turning off residual tuning (i.e., MCP++ → MCP) causes only a minor drop, whereas the same ablation on DPE leads to a much larger degradation showing that  that our initial prototypes are intrinsically stronger. Moreover, DPE leverages only low‑entropy samples and lacks a negative cache to constrain and calibrate its visual prototypes. Furthermore, our method does not adopt the textual prototype evolution mechanism used in DPE.

\subsection{PromptAlign \cite{PromptAlign}}
PromptAlign~\cite{PromptAlign} is the only existing method that explicitly considers the distribution of test samples. However, it still differs significantly from our MCP approach. Specifically, PromptAlign relies on an ImageNet subset as the source domain and computes the mean and variance of the token representations produced by the CLIP encoder on this subset. At test time, it constrains the prompt‐tuning process by aligning the token statistics of test samples to the source‐domain statistics, thereby reducing the domain shift between source and test domains. In contrast, MCP requires no ImageNet subset for initialization; instead, it dynamically constructs the Prototype Center at test time by combining textual and visual information and uses this center solely for sample selection rather than as part of the optimization loss. Moreover, MCP does not access internal token‐level statistics (such as mean and variance) of the encoder, but only retrieves the final [CLS] embedding and applies a lightweight residual vector for fine‐tuning.

\label{sec:additional_results}
\begin{table*}[!h]
    \renewcommand{\arraystretch}{1.2}  
    \setlength{\tabcolsep}{5pt}       
    \fontsize{9}{8.5}\selectfont
    \centering
    \caption{Results on the Cross-Domain Benchmark. Top-1 accuracy (\%) results are presented for all evaluated methods employing the
ResNet-50 visual backbone of CLIP. The best results are highlighted in \textbf{bold}.}
    \label{tab:performance_comparison_cs}
    \begin{tabular}{l|cccccccccc|c}
    \toprule
    \multicolumn{1}{l}{\textbf{Method}} & 
    \multicolumn{1}{c}{\textbf{Aircraft}} & 
    \multicolumn{1}{c}{\textbf{Caltech}} & 
    \multicolumn{1}{c}{\textbf{Cars}} & 
    \multicolumn{1}{c}{\textbf{DTD}} & 
    \multicolumn{1}{c}{\textbf{EuroSAT}} & 
    \multicolumn{1}{c}{\textbf{Flower}} & 
    \multicolumn{1}{c}{\textbf{Food101}} & 
    \multicolumn{1}{c}{\textbf{Pets}} & 
    \multicolumn{1}{c}{\textbf{SUN397}} & 
    \multicolumn{1}{c}{\textbf{UCF101}} & 
    \multicolumn{1}{c}{\textbf{Average}} \\
    \midrule
    CLIP--ResNet50 & 15.66 & 85.88 & 55.70 & 40.37 & 23.69 & 61.75 & 73.97 & 82.57 & 58.80 & 58.84 & 55.82 \\
    \rowcolor{mygray}CoOp & 15.12 & 86.53 & 55.32 & 37.29 & 26.20 & 61.55 & 75.59 & 87.00 & 58.15 & 59.05 & 56.18 \\
    \midrule
    \rowcolor{mypink2}TPT & 17.58 & 87.02 & 58.46 & 40.84 & 28.33 & 62.69 & 74.88 &84.49 & 61.46 & 60.82 & 57.66 \\
    \rowcolor{mypink2}DiffTPT & 17.60 & 86.89 & 60.71 & 40.72 & 41.04 & 63.53 & \textbf{79.21}& 83.40 & 62.72 & 62.67 & 59.85 \\
    \midrule
    \rowcolor{mypink1} TDA & 17.61 & 89.70 & 57.78 & 43.74 & 42.11 & 68.74 & 77.75 & 86.18 & 62.53 & 64.18 & 61.03 \\
    \rowcolor{mypink1} DMN & 22.77 & 90.14 & 60.02 & 50.41 & 48.72 & 67.93 & 76.70 & 86.78 & 64.39 & 65.34 & 63.71 \\
    \rowcolor{mypink1} DPE & 19.80 & 90.83 & 59.26 & 50.18 & 41.67 & 67.60 & 77.83 & 85.97 & 64.23 & 61.98 & 61.93\\

    \midrule
    \rowcolor{mypink} \textbf{MCP}   & 23.04 & 90.99 & 61.27 & 53.19 & 55.60 & 68.49 & 78.37 & 87.35 & 65.39 & 67.14 & 65.08\\
    \rowcolor{mypink} \textbf{MCP++} & \textbf{23.40} & \textbf{91.13} & \textbf{61.76} & \textbf{53.61} & \textbf{55.74} & \textbf{69.96} & 78.44 & \textbf{87.49} & \textbf{65.55} & \textbf{67.86} & \textbf{65.49}\\
    \bottomrule
    \end{tabular}
\end{table*}

\begin{table*}[!h]
    \renewcommand{\arraystretch}{1.2}  
    \setlength{\tabcolsep}{8.5pt}       
    \fontsize{9}{8.5}\selectfont
    \centering
    \caption{Results on the OOD Benchmark. Top-1 accuracy (\%) results are presented for all evaluated methods employing the ViT-B/16
visual backbone of ResNet-50. The best results are highlighted in \textbf{bold}}
    \label{tab:performance_comparison_ood}
    \begin{tabular}{l|ccccc|c|c}
    \toprule
    \multicolumn{1}{l}{\textbf{Method}} & 
    \multicolumn{1}{c}{\textbf{ImageNet}} & 
    \multicolumn{1}{c}{\textbf{A}} & 
    \multicolumn{1}{c}{\textbf{V2}} & 
    \multicolumn{1}{c}{\textbf{R}} & 
    \multicolumn{1}{c}{\textbf{S}} & 
    \multicolumn{1}{c}{\textbf{OOD Avg.}} & 
    \multicolumn{1}{c}{\textbf{Average}} \\
    \midrule
    CLIP--ResNet50     & 58.16 & 21.83 & 51.41 & 56.15 & 33.37 & 40.69 & 44.18 \\
    \rowcolor{mygray}CoOp       & 63.33 & 23.06 & 55.40 & 56.60 & 34.67 & 42.43 & 46.61 \\
    \midrule
    \rowcolor{mypink2}TPT        & 60.74 & 26.67 & 54.70 & 59.11 & 35.09 & 43.89 & 47.26 \\
    \rowcolor{mypink2}DiffTPT    & 60.80 & \textbf{31.06} & 55.80 & 58.80 & 37.10 & 45.69 & 48.71 \\
    \midrule
    \rowcolor{mypink1} TDA       & 61.35 & 30.29 & 55.54 & 62.58 & 38.12 & 46.63 & 49.58 \\
    \rowcolor{mypink1} DMN       & 63.87 & 28.57 & 56.12 & 61.44 & 39.84 & 46.49 & 49.97 \\
    \rowcolor{mypink1} DPE       & 63.41 & 30.15 & 56.72 & 63.72 & 40.03 & 47.66 & 50.81 \\
    \midrule
    \rowcolor{mypink} \textbf{MCP}   & 64.19 & 29.30 & 56.93 & 64.40 & 41.16 & 48.02 & 51.20 \\
    \rowcolor{mypink} \textbf{MCP++} & \textbf{64.44} & 29.29 & \textbf{57.12} & \textbf{64.92} & \textbf{41.35} & \textbf{48.17} & \textbf{51.42} \\
    \bottomrule
    \end{tabular}
\end{table*}

\subsection{Prototype Learning}
Prototype learning was formally introduced to the few‑shot classification paradigm by Snell \emph{et al.} \cite{snell2017prototypical} and has since rapidly expanded into a wide array of research domains: semantic segmentation \cite{shaban2017oneshot,liu2025probabilistic}, object detection \cite{Kang2018FewShotOD}, OOD detection \cite{li2025dpu, chen2024proto, lu2024learning}, continual learning \cite{rebuffi2017icarl,gou2025queryable,fukuda2025adapter}, action recognition \cite{ni2022multimodal}, domain adaption \cite{zhang2024node, clusteradapter,TPS,DPE,Yu2022TaskRF,zhu2025dynamic,zhang2023unsupervised,palanisamy2024proto,ali2025dpa,liang2025advancing,qu2025learning}. Similar to previous work \cite{TPS,DPE,clusteradapter,Yu2022TaskRF}, we construct text prototypes by averaging the embeddings of multiple prompts and derive visual prototypes by averaging the cached image features. Because plain averaging often yields prototypes that are neither discriminative nor robust, much recent research has focused on refining or calibrating prototype quality. Representative approaches include: DPE \cite{DPE}, which co‑evolves visual and textual prototypes while learning a multimodal residual online for precise alignment; ClusterAdapter \cite{clusteradapter}, which clusters prototypes and performs fine‑grained tuning inside a lightweight adapter; TPS \cite{TPS}, which combines richer textual descriptions with online residual learning to correct text prototypes; BPRE \cite{qiao2025bidirectional}, which introduces a multidimensional quality‑aware reward and a prototype–reward interactive evolution mechanism; and ProtoMM \cite{zhu2025dynamic} aligns visual and textual particles via optimal transport and dynamically re‑weights them to refine multimodal prototypes. Our method MCP boosts visual prototype quality \textbf{without extra training} by screening test samples on two criteria, entropy and distance to the prototype center. The extended variant, MCP++, performs prototype residual tuning at test time, incorporating cross-modal alignment and negative-sample distance constraints to further improve prototype accuracy and robustness.

\section{Additional Experimental Results}

\subsection{Cross-Dataset Generalization} Table~\ref{tab:performance_comparison_cs} presents the results of our cross-dataset generalization experiments on the ResNet-50 backbone, comparing our method against state-of-the-art approaches across ten diverse datasets. The results demonstrate that our proposed methods, MCP and MCP++, achieve superior performance under significant distributional shifts. On the ResNet-50 backbone, our methods achieve the best performance on 9 out of 10 tasks, with MCP++ obtaining the highest average accuracy of 65.49\%. Compared to both prompt-based methods (e.g., CoOp~\cite{CoOp}, TPT~\cite{TPT}, and DiffTPT~\cite{DiffTPT}) and cache-based methods (e.g., TDA~\cite{TDA}, DPE~\cite{DPE}, DMN~\cite{Zhang2024DualMN}), MCP and MCP++ achieve superior results.

\subsection{Natural Distribution Shifts} We evaluate the robustness of our proposed methods on in-domain ImageNet and its four out-of-distribution variants using the ResNet-50 backbone, with the results presented in Table~\ref{tab:performance_comparison_ood}. Our approach outperforms leading prompt-tuning methods, achieving a 4.16\% improvement over TPT with MCP++ and a 2.71\% improvement over DiffTPT. When compared to cache-based methods, MCP++ surpasses TDA by 1.84\%, DMN by 1.45\%, and DPE by 0.61\%. This result indicates that our method is generally effective in both domain-specific variations and out-of-distribution robustness scenarios, demonstrating its ability to maintain high performance under natural distribution shifts.

\subsection{Ablation Study Results for Losses} As shown in Table \ref{tab:ablation}, introducing either the alignment or contrastive loss improves model performance compared to the baseline. Specifically, applying only the alignment loss yields 30.07\% accuracy, while the contrastive loss alone achieves 29.92\%.
Notably, combining both losses leads to the best result, with 30.18\% accuracy on the Aircraft dataset.
\begin{table}[!h]
    \centering
    \renewcommand{\arraystretch}{1.2}  
    \setlength{\tabcolsep}{8.5pt}        
    \fontsize{8.5}{8.5}\selectfont
    \caption{Ablation Study Results for Align and Contrast Losses}
    \label{tab:ablation}
    \begin{tabular}{cc|c}
    \toprule
    \multicolumn{1}{c}{\textbf{$L_{\text{align}}$}} & 
    \multicolumn{1}{c}{\textbf{$L_{\text{contrast}}$}} & 
    \multicolumn{1}{c}{\textbf{Aircraft}} \\
    \midrule
     &  & 29.74 \\
    \checkmark &  & 30.07 \\
     & \checkmark & 29.92 \\
    \checkmark & \checkmark & \textbf{30.18} \\
    \bottomrule
    \end{tabular}
\end{table}

\subsection{More Sensitivity Analyses of Cache Size}
In Fig.4 (Middle), we have analyzed the sensitivity of model performance to the sizes of the Align Cache and Entropy Cache on cross-domain datasets. We further investigate the effect of the Negative Cache size on the model’s adaptation capability.

As shown in Table \ref{tab:neg_cache_size}, the model achieves optimal performance when the negative cache size is set to a moderate value (e.g., 3), while both smaller and larger cache sizes lead to performance degradation. This observation suggests that a moderate number of high-entropy negative samples helps mitigate the impact of noisy pseudo-labels and suppress misclassification near category boundaries. However, an excessive number of negative samples may introduce additional noise, impairing the model’s discriminative ability.

Overall, model performance exhibits a certain degree of sensitivity to cache size. Larger cache capacity does not necessarily yield better results; instead, a balance must be struck between “information coverage” and “representation purity.” In addition, we observe that the optimal cache configuration may vary across datasets. To ensure consistency and fairness throughout all experiments, we adopt a unified configuration of $\lvert M_c^{\text{entropy}} \rvert = 10$, $\lvert M_c^{\text{align}} \rvert = 10$, and $\lvert M_c^{\text{negative}} \rvert = 3$.

\begin{table}[!h]
    \centering
    \renewcommand{\arraystretch}{1.2}  
    \setlength{\tabcolsep}{10pt}      
    \fontsize{9}{9}\selectfont
    \caption{Effects of different negative cache sizes on cross-domain datasets under ResNet-50.}
    \label{tab:neg_cache_size}
    \begin{tabular}{c|ccc}
        \toprule
        \textbf{Negative Cache Size} & 2 & 3 & 5 \\
        \midrule
        \textbf{Accuracy (\%)} & 64.97 & \textbf{65.08} & 64.73 \\
        \bottomrule
    \end{tabular}
\end{table}

\section{Theoretical Analysis}

In this section, we demonstrate that incorporating Align Cache further reduces the error upper bound compared to using only Entropy Cache. First, we present the fundamental theoretical setup for our analysis; then, we introduce several key assumptions. Finally, we derive the error upper bounds for both approaches and prove that the method employing Align Cache achieves a tighter bound. We adopt the proof strategy of BoostAdapter \cite{Zhang2024BoostAdapterIV}. Unlike their setting, which automatically satisfies the Strong Density Condition in Assumption~\ref{sec:Assumptions} (i.e., \(c_a > c_t\)) due to its construction, our setting does not guarantee this inequality \emph{a priori}. Therefore, we explicitly prove that \(c_a > c_t\) holds within our framework.
\subsection{Problem Setting.} 
\label{sec:setting}

We formalize the foundational setup for test-time adaptation. Consider a binary classification task (extensible to multi-class scenarios) with joint data distribution \( p_t(x,y) \) over the target domain. Assume we observe \( n \) i.i.d. test samples:
\[
\{(x_i,y_i)\}_{i=1}^n \stackrel{\text{i.i.d.}}{\sim} p_t(x,y)
\]
where \( y_i \in \{0,1\} \) represents one-hot encoded labels in the binary case (and, in the multi-class case, \( y_i \) is an one-hot vector of length \( N \)).

\subsection{Definitions} 

\paragraph{Classification Error.}
For a binary classifier \( f: \mathcal{X} \to \{0,1\} \), its error under distribution \( p_t(x,y) \) is defined as:
\begin{equation}
\epsilon(f) = \mathbb{E}_{p_t(x,y)}[\mathbf{1}_{\{f(x) \neq y\}}] = \mathbb{E}_{p_t(x,y)}[|f(x) - y|]
\end{equation}
The equality holds specifically for binary classification.

\paragraph{Excess Error.}
Let \( f^* \) denote the Bayes-optimal classifier defined as 
\[
f^*(x) = \mathbb{I}_{\{\eta(x) \geq 1/2\}}
\]
where \( \eta(x) = \mathbb{E}[y|x] \). The excess error of \( f \) is:
\begin{equation}
\mathcal{E}(f) = \epsilon(f) - \epsilon(f^*) = 2\mathbb{E}_{x \sim p_t(x)}\left[\left|\eta(x)-\frac{1}{2}\right| \cdot \mathbb{I}_{\{f(x) \neq f^*(x)\}}\right]
\end{equation}

\paragraph{Cache Classifier.}
Given an encoder \(g: \mathcal{X} \to \mathbb{R}^d\) and a cache of \(K\) stored samples, Tip-Adapter~\cite{TipAdapter} proposes
\begin{equation}
    \label{equ:cache}
    \boldsymbol{p}_{\text{cache}}(x)
    \;=\;
    A\bigl(g(x)\,G_{\text{cache}}^\top\bigr)\,Y,
\end{equation}
where \(A(z) = \alpha \exp\bigl(-\beta(1-z)\bigr)\) is a scaling function, \(G_{\text{cache}} \in \mathbb{R}^{K\times d}\) contains features of cached samples, and \(Y \in \mathbb{R}^{K\times N}\) holds their labels.

\paragraph{Aligned Distribution \texorpdfstring{\boldmath $p_a(x)$}{p_a(x)}}
In addition to the original target distribution \(p_t(x,y)\), we define an aligned distribution \(p_a(x)\) that focuses on test samples near a category anchor \(\mu_c\in \mathbb{R}^d\). Specifically, we retain only those \(x\) satisfying
\[
\|g(x) - \mu_c\| \;\le\; d_0,
\]
where \(d_0>0\) is a constraint threshold. Formally,
\[
p_a(x)
\;=\;
p_t\!\bigl(x\mid \|g(x) - \mu_c\|\le d_0\bigr).
\]
Hence, \(p_a(x)\) captures the center-constrained subset of the target domain.

\paragraph{Aligned Cache Classifier.}
While existing cache-based methods generally keep only historical test samples, we further incorporate \emph{aligned samples} near \(\mu_c\). Suppose \(k_t\) historical samples and \(k_a\) aligned samples reside in the cache. Building on Eq.~\ref{equ:cache}, we define:
\begin{equation}
    \label{equ:aligncache}
    \boldsymbol{p_{\text{align}}}(x) 
    \;=\;
    A\bigl(g(x)\,\tilde{G}_{\text{cache}}^\top\bigr)\,\tilde{Y},
\end{equation}
where \(\tilde{G}_{\text{cache}} \in \mathbb{R}^{(k_t+k_a)\times d}\) mixes historical and aligned features, and \(\tilde{Y}\in \mathbb{R}^{(k_t+k_a)\times N}\) is the corresponding label matrix.

\subsection{Practical Implementation.}
In practice, we adopt an entropy-based threshold to pick reliable historical samples, and require 
$\|g(x) - \mu_c\| \le d_0$ for the aligned ones. The distance \(d_0\) is computed dynamically from the cache statistics, defined as the distance between the class anchor \(\mu_c\) and the farthest retained sample in the cache.
 Each class is capped at $k$ slots, replacing higher-entropy or out-of-reach samples. We omit further details here, as our main focus is on the theoretical guarantees.

\subsection{Assumptions}
\label{sec:Assumptions}
\paragraph{Strong Density Condition.}
Let \(x_0\) be any test sample from \(p_t(x)\) and also from the aligned distribution \(p_a(x_0)\). We assume there exist positive constants \(m\) and \(M\), scaling factors \(c_t\) and \(c_a\), and a radius \(R>0\). Define \(\mathcal{B}(x,r)=\{x'\mid\|x'-x\|\le r\}\). We assume \(p_t(x)\) and \(p_a(x_0)\) are absolutely continuous w.r.t.\ the Lebesgue measure \(\lambda\) in \(\mathbb{R}^d\). Then, for any \(r\in(0,R]\),
\[
\left\{
\begin{aligned}
&\lambda\bigl[p_t(x)\cap \mathcal{B}(x_0,r)\bigr]
\;\ge\;
c_t \,\lambda\bigl[\mathcal{B}(x_0,r)\bigr],\\[3pt]
&\lambda\bigl[p_a(x_0)\cap \mathcal{B}(x_0,r)\bigr]
\;\ge\;
c_a \,\lambda\bigl[\mathcal{B}(x_0,r)\bigr],\\[3pt]
&m
\;<\;\tfrac{dp_t(x)}{d\lambda}
\;<\;M,\quad
m
\;<\;\tfrac{dp_a(x)}{d\lambda}
\;<\;M.
\end{aligned}
\right.
\]
Intuitively, in any local region, the target distribution occupies at least \(c_t\) fraction of that area's mass, while the aligned distribution $p_a$ has an area fraction \(c_a\) that is strictly greater than \(c_t\). 

\medskip
\noindent
\textbf{Proof of \(c_a > c_t\).} 
Recall that we defined the aligned distribution as 
\[
p_a(x) = p_t\Bigl(x\mid \|g(x)-\mu_c\|\le d_0\Bigr),
\]
which, by the definition of conditional probability, can be written as
\[
p_a(x) = \frac{p_t(x)}{P(\|g(x)-\mu_c\|\le d_0)},\quad \text{for } \|g(x)-\mu_c\|\le d_0.
\]
Now, consider any test sample \(x_0\) and any ball \( \mathcal{B}(x_0,r)\) (with \(r\in (0,R]\)) that is fully contained in the region 
\[
\{x:\|g(x)-\mu_c\|\le d_0\}.
\]
For the aligned distribution, the probability mass within \(\mathcal{B}(x_0,r)\) is given by
\begin{equation}
\begin{split}
\lambda\Bigl[p_a(x)\cap \mathcal{B}(x_0,r)\Bigr] 
&= \int_{\mathcal{B}(x_0,r)} p_a(x) \, dx \\
&= \frac{1}{P(\|g(x)-\mu_c\|\le d_0)} 
\int_{\mathcal{B}(x_0,r)} p_t(x) \, dx.
\end{split}
\end{equation}
By the Strong Density Condition for the target distribution, we have
\[
\int_{\mathcal{B}(x_0,r)} p_t(x) \, dx \ge c_t\,\lambda\Bigl[\mathcal{B}(x_0,r)\Bigr].
\]
Thus,
\[
\lambda\Bigl[p_a(x)\cap \mathcal{B}(x_0,r)\Bigr] \ge \frac{c_t}{P(\|g(x)-\mu_c\|\le d_0)}\,\lambda\Bigl[\mathcal{B}(x_0,r)\Bigr].
\]
By definition of the local density constant for the aligned distribution, we require that
\[
\lambda\Bigl[p_a(x)\cap \mathcal{B}(x_0,r)\Bigr] \ge c_a\,\lambda\Bigl[\mathcal{B}(x_0,r)\Bigr].
\]
Comparing the two inequalities, it follows that
\[
c_a \ge \frac{c_t}{P(\|g(x)-\mu_c\|\le d_0)}.
\]
Since the filtering condition ensures that not all samples satisfy \(\|g(x)-\mu_c\|\le d_0\), we have 
\[
P(\|g(x)-\mu_c\|\le d_0) < 1.
\]
Therefore, it must hold that
\[
c_a > c_t.
\]

\medskip
\noindent
\textbf{L-Lipschitz Condition.}
We assume there is a positive constant \(L\) such that the classification function \(f\) is \(L\)-Lipschitz:
\[
|f(x)-f(x')|\;\le\;L\,\|x-x'\|.
\]
Intuitively, $f$ is smooth, so small changes in input cannot produce large output variations.

\medskip
\noindent
\textbf{Low Noise Condition.}
Let \(\beta\) and \(C_\beta\) be positive constants. We assume that for any \(t>0\),
\[
P_{x\sim p_t(x)}\!\Bigl(\bigl|f(x)-\tfrac12\bigr|<t\Bigr)\;\le\;C_\beta\,t^\beta.
\]
Intuitively, the probability mass near the threshold ($f(x)\approx\tfrac12$) is small, ensuring high confidence in that region.

\subsection{Proof Sketch}
Under these three assumptions, we show how the cache classifier can achieve low empirical risk by retrieving sufficiently representative samples. The key arguments rely on:
\begin{itemize}
    \item \textbf{Strong Density:} guarantees local coverage of the target domain and (separately) of the aligned distribution near \(\mu_c\). In particular, $c_a > c_t$ implies the aligned region is “denser” when restricted to center-based samples.
    \item \textbf{Lipschitz Smoothness:} ensures $f$ does not fluctuate excessively around similar points.
    \item \textbf{Low Noise:} limits the mass near the decision boundary, avoiding high uncertainty.
\end{itemize}
We leverage these properties to analyze two scenarios: one that uses only an entropy cache, and another that combines it with an align cache.

\subsection{Propositions}

Based on the above assumptions, we propose the following two propositions. These results were originally derived in AdaNPC\cite{Zhang2023AdaNPCEN} and further extended in Boostadapter\cite{Zhang2024BoostAdapterIV}.
For more detailed derivations and complete proofs, please refer to the appendices of both AdaNPC\cite{Zhang2023AdaNPCEN} and Boostadapter\cite{Zhang2024BoostAdapterIV}.

\medskip
\noindent
\textbf{Proposition 1 (Entropy Cache Reduces Empirical Risk).}
Consider a training-free classifier \(f\) that uses only historical low-entropy samples, as defined in Eq.~\ref{equ:cache}. Let \(n_t\) be the number of confidently predicted low-entropy samples from the target domain, and \(k_t\) the subset of those stored in the cache. Under Assumptions 1--3, and for sufficiently large \(n_t\) and \(k_t\), it holds with high probability that
\begin{equation}
\label{prop:his_only}
\mathcal{E}(f) \;\le\;
\mathcal{O}\!\Bigl(
\bigl(\tfrac{1}{k_t}\bigr)^{\tfrac14}
\;+\;
\bigl(\tfrac{k_t}{c_t\,n_t}\bigr)^{\tfrac{1}{d}}
\Bigr)^{\,1+\beta}.
\end{equation}
This guarantees that by selecting \(k_t\) high-quality historical samples out of \(n_t\), one can achieve low empirical risk if \(n_t\) is large enough, since the cache captures representative information from the target domain.

\medskip
\noindent
\textbf{Proposition 2 (Entropy Cache Benefits from Align Cache).}
Next, let \(n_t\) again be the count of low-entropy historical samples, while \(n_a\) is the number of aligned samples obtained via distance-to-center filtering. Suppose the cache stores \(k_t\) low-entropy samples and \(k_a\) aligned samples. Let $ w_{ti} $ and $ w_{ai} $ denote the instance weights assigned to low-entropy and aligned samples, respectively, with the explicit normalization constraint $\sum_{i=1}^{k_t} w_{ti} + \sum_{i=1}^{k_a} w_{ai} = 1$.
 Then, the empirical risk of \(f\) (as in Eq.~\ref{equ:aligncache}) is bounded by
\begin{equation}
\label{prop:his_plus_align}
\begin{aligned}
\mathcal{E}(f)\;\le\;\;
\mathcal{O}\Bigl(
&\bigl(\tfrac{1}{k_t + k_a}\bigr)^{\tfrac{1}{4}}
\;+\;
\sum_{i=1}^{k_t} w_{ti}\,\bigl(\tfrac{k_t}{c_t\,n_t}\bigr)^{\tfrac{1}{d}}\\
&+\;
\sum_{i=1}^{k_a} w_{ai}\,\bigl(\tfrac{k_a}{c_a\,n_a}\bigr)^{\tfrac{1}{d}}
\Bigr)^{\,1+\beta}.
\end{aligned}
\end{equation}
Hence, the historical cache can reduce empirical risk even further by incorporating \(k_a\) aligned samples, thereby capturing both general historical information and center-focused aligned information about the target domain.

\subsection{Proof of Proposition 2 Improvement}

We now show that, incorporating aligned samples leads to a strictly tighter error bound than using only historical samples.
For a more rigorous proof, please refer to the appendices of AdaNPC~\cite{Zhang2023AdaNPCEN} and BoostAdapter~\cite{Zhang2024BoostAdapterIV}.

\medskip
\noindent
\textbf{Step 1: Retention Ratio Consistency} \\
Let the filtering rule select samples satisfying event $ A $ (Here, entropy is less than $\leq \tau$), which is \textit{statistically independent} of the alignment condition $\mathcal{R} = \{x : \|g(x)-\mu_c\| \leq d_0\}$. Then:

\begin{itemize}
\item For historical samples from $ p_t(x) $:
\[
\frac{k_t}{n_t} = p_t(A) 
\]

\item For aligned samples from $ p_a(x) = p_t(x \mid \mathcal{R}) $:
\[
\frac{k_a}{n_a} = p_t(A \mid \mathcal{R}) \overset{\text{(indep.)}}{=} p_t(A) 
\]
\end{itemize}

\medskip
\noindent
This implies:
\begin{equation}
\frac{k_t}{n_t} = \frac{k_a}{n_a} \quad 
\end{equation}

\medskip
\noindent
\textbf{Step 2: Error Bound Reformulation} \\
Since we have normalized weights $ w_{ti}, w_{ai} $ as:
\[
\sum_{i=1}^{k_t} w_{ti} + \sum_{i=1}^{k_a} w_{ai} = 1
\]
\medskip
\noindent
Let $ a = \sum_{i=1}^{k_t} w_{ti} $, $ 1-a = \sum_{i=1}^{k_a} w_{ai} $. The bounds become:

\begin{itemize}
\item Proposition 1 (Historical only):
\[
\mathcal{E}(f) \leq \mathcal{O}\Biggl( 
\,\underbrace{\biggl(\frac{1}{k_t}\biggr)^{\!\tfrac{1}{4}}}_{\text{\rm Term 1}} 
\,+\,
\underbrace{\biggl(\frac{k_t}{c_t n_t}\biggr)^{\!\tfrac{1}{d}}}_{\text{\rm Term 2}}\,
\Biggr)^{\!\!1+\beta}
\]

\item Proposition 2 (Historical + Aligned):
\[
\begin{aligned}
\mathcal{E}(f)\;\le\;\;
\mathcal{O}\Biggl(
&\underbrace{\biggl(\frac{1}{k_t + k_a}\biggr)^{\!\tfrac{1}{4}}}_{\text{\rm Term' 1}} 
\\
&\quad + \underbrace{\biggl(
a\,\Bigl(\frac{k_t}{c_t n_t}\Bigr)^{\!\tfrac{1}{d}} 
+ (1{-}a)\Bigl(\frac{k_a}{c_a n_a}\Bigr)^{\!\tfrac{1}{d}}
\biggr)}_{\text{\rm Term' 2}} 
\Biggr)^{\!\!1+\beta}
\end{aligned}
\]
\end{itemize}

\medskip
\noindent
\textbf{Step 3: Strict Dominance of Proposition 2}

\begin{itemize}
\item
{Sample Size Advantage}
\[
\underbrace{\left(\frac{1}{k_t + k_a}\right)^{\frac{1}{4}}}_{\text{Term' 1}}
\;<\;
\underbrace{\left(\frac{1}{k_t}\right)^{\frac{1}{4}}}_{\text{Term 1}}
\quad (\because\, k_t + k_a > k_t).
\]
\item 
{Convex Combination Advantage}
From 
\[
\frac{k_t}{n_t} \;=\; \frac{k_a}{n_a}
\quad \text{and} \quad
c_a > c_t:
\]

\[
\frac{k_a}{c_a n_a}
\;=\;
\frac{k_t}{c_a n_t}
\;<\;
\frac{k_t}{c_t n_t}
\;\Longrightarrow\;
\left(\frac{k_a}{c_a n_a}\right)^{\frac{1}{d}}
\;<\;
\left(\frac{k_t}{c_t n_t}\right)^{\frac{1}{d}}.
\]

Thus,
\[
\underbrace{a \left(\frac{k_t}{c_t n_t}\right)^{\frac{1}{d}}
\;+\;(1-a)\left(\frac{k_a}{c_a n_a}\right)^{\frac{1}{d}}}_{\text{Term' 2}}
\;<\;
\underbrace{\left(\frac{k_t}{c_t n_t}\right)^{\frac{1}{d}}}_{\text{Term 2}}.
\]

\item {Combined Effect}
\[
\underbrace{\text{Term' 1 + Term' 2}}_{\text{Proposition 2}}
\;<\;
\underbrace{\text{Term 1 + Term 2}}_{\text{Proposition 1}}.
\]

\end{itemize}

\subsection{Concluding Discussion}
Overall, we have theoretically explored the problem of test-time adaptation by developing a cache-based classifier model and defining the relevant concepts. Under the assumptions of strong density, Lipschitz smoothness, and low noise, we analyzed the error bounds for both using only historical samples and combining them with aligned samples. The theoretical proofs demonstrate that the overall error bound is significantly smaller when aligned samples are incorporated, providing solid theoretical support for the introduction of the Align Cache method.



\end{document}